\documentclass[acmsmall,screen]{acmart}
\usepackage{booktabs}
\usepackage{tabularx}
\usepackage{xurl}
\usepackage[most]{tcolorbox}
\usepackage{pifont}
\usepackage{url}
\usepackage{xspace}
\usepackage{subcaption}
\usepackage{needspace}

\newcommand{\tLead}{\ensuremath{T_{\mathrm{lead}}}\xspace}
\newcommand{\tRequired}{\ensuremath{T_{\mathrm{required}}}\xspace}
\newcommand{\tEarliest}{\ensuremath{T_{\mathrm{earliest}}}\xspace}
\newcommand{\appFr}[1]{\ensuremath{\mathtt{IT_{f}^{#1}}}\xspace}
\newcommand{\appOd}[1]{\ensuremath{\mathtt{IT_{fol}^{#1}}}\xspace}

\newcommand{\answerRQ}[1]{\begin{tcolorbox}[size=title, colframe=black, width=1\linewidth, colback=gray!5, breakable]#1\end{tcolorbox}}

\AtBeginDocument{%
}

\setcopyright{acmlicensed}
\copyrightyear{2026}
\acmYear{2026}
\acmDOI{XXXXXXX.XXXXXXX}

\acmJournal{TOSEM}
\acmVolume{37}
\acmNumber{4}
\acmArticle{111}
\copyrightyear{2026}
\acmYear{2026}
\acmMonth{11}

\begin{document}

\title{Assessing Vision-Language Models for Early Prediction of Unsafe Situations in Robotics}

\author{Muhammad Farooq}
\email{muhammad.farooq@oslomet.no}
\orcid{0009-0002-5774-082X}
\affiliation{%
\institution{OsloMet -- Oslo Metropolitan University}
\city{Oslo}
\country{Norway}
}

\author{Aitor Arrieta}
\email{aarrieta@mondragon.edu}
\orcid{0000-0001-7507-5080}
\affiliation{%
\institution{Mondragon Unibertsitatea}
\city{Mondragon}
\country{Spain}
}

\author{Paolo Arcaini}
\email{paolo@simula.no}
\orcid{0000-0002-6253-4062}
\affiliation{%
\institution{Simula Research Laboratory}
\city{Oslo}
\country{Norway}
}

\author{Shaukat Ali}
\email{shaukat@simula.no}
\orcid{0000-0002-9979-3519}
\affiliation{%
\institution{Simula Research Laboratory}
\city{Oslo}
\country{Norway}
}
\affiliation{%
\institution{OsloMet -- Oslo Metropolitan University}
\city{Oslo}
\country{Norway}
}
\author{Thomas Peyrucain}
\email{thomas.peyrucain@pal-robotics.com}
\orcid{0009-0005-2500-570X}
\affiliation{%
\institution{PAL Robotics}
\city{Barcelona}
\country{Spain}
}

\renewcommand{\shortauthors}{Farooq et al.}

\begin{abstract}
Vision-Language Models (VLMs) process visual information and textual instructions and, by using their ability to understand and reason about context, are well-suited for predicting robot failures before they occur. However, evidence supporting this remains limited. To address this, we present an empirical evaluation of VLMs for predicting robot failures. In particular, we experiment with Qwen, InternVL, and LLaVA-OneVision across three robotic case studies (HuRoN, Pal Robotics, and CrowdBot), analyzing both video and odometry data for early failure warning (i.e., before an unsafe situation occurs). Specifically, we evaluate their performance in classifying situations as safe, potentially unsafe, and unsafe, together with uncertainty quantified using various metrics, and study the relationship between uncertainty and performance. Our results showed that Qwen provides the strongest overall classification and prediction-reliability performance, while InternVL shows stronger early-warning capability in several configurations. The results further show that the effect of robot-state information and temporal history depends on the model and evaluation objective. Overall, our findings indicate that general-purpose VLMs can support early robot-safety monitoring, but the most suitable configuration depends on whether the priority is accurate assessment of safety-state, reliable confidence and uncertainty, or timely warning before unsafe events. 
\end{abstract}




\maketitle

\section{Introduction}
Autonomous navigation robots are often deployed in environments with humans and other obstacles, such as offices, warehouses, and other crowded places, where they must perform their intended tasks correctly while also avoiding collisions with people and other objects. Consequently, they must be able to detect potential failures early and prevent them. Very often, the software in these robots includes functionality for object detection and failure prediction, typically using deep learning models~\cite{technologies12090157}. However, deep learning models often perform well only on the types of data they are trained on, and their performance degrades when they encounter unseen objects~\cite{9764831}.

To address the above limitation, foundation models such as Vision-Language Models (VLMs) are often advocated for their ability to generalize, enabling them to handle unseen situations~\cite{yousaf2026vlmUnderwater}. To enable robots to detect failures in advance and issue warnings about potential failures, VLMs could provide an effective solution; however, evidence supporting their use in this context remains limited. VLMs, however, have been used for various applications in robotics. For example, VLMs have been used to generate test scenarios for robotic applications~\cite{wu2026vlmTestingAMR} and to automate test oracles~\cite{10.1145/3805760.3814910}. In addition, VLMs and related embodied multimodal models have been explored for robotic reasoning, task planning, and control~\cite{DBLP:journals/corr/abs-2303-03378,DBLP:journals/corr/abs-2307-15818}. Recent video-based VLM navigation studies have further shown that temporal visual observations can provide useful context for navigation decisions in unseen environments~\cite{Zhang-RSS-24}. Nevertheless, existing studies primarily focus on task execution, navigation planning, and control, leaving the potential of VLMs to predict robot safety issues from historical observations insufficiently explored.

To address this gap, we make the following contributions. We provide a systematic empirical evaluation of three VLMs---Qwen, InternVL, and LLaVA-OneVision---for robot-safety classification across three robotic case studies. We compare two input types: Visual History Only \appFr{} and visual history enriched with odometry and LiDAR information\appOd{}. For both input types, we evaluate how the length of the visual history affects classification performance using different history settings. We go beyond classification performance by evaluating the confidence, uncertainty, and calibration of VLM predictions, providing evidence on how reliable their predictions are for safety-critical use. Finally, we evaluate VLMs as early-warning mechanisms at the event level, distinguishing actionable, too-early, too-late, and missed warnings. This connects classification performance with the practical question of whether a model can identify an unsafe situation early enough for a robot or its digital twin to take preventive action.

Results showed that the evaluated VLMs differ substantially in their suitability for robot-safety monitoring. Qwen provides the strongest overall performance for safety classification and also shows the most reliable confidence, uncertainty, and calibration behavior. The results further indicate that adding robot-state information such as odometry and LiDAR can improve class-balanced safety prediction, while the effect of temporal history depends on the model and input setting. Importantly, the model that performs best for classifying the current safety state is not always the one that provides the earliest or most actionable warning. Overall, the findings show that general-purpose VLMs have promising potential for early robot-safety monitoring, but their configuration should be selected according to whether the main objective is accurate state assessment, reliable prediction confidence, or timely warning. 


\section{Related Work}
\label{sec:relatedWork}
We review prior work from three perspectives that directly support our study: robot perception and navigation, VLM-based failure monitoring, and prediction reliability. Together, these areas provide the context for evaluating VLMs as early-warning components in robot digital twins.
\subsection{Learning-Based Perception and VLMs in Robotics}
\label{subsec:P1}
Autonomous Mobile Robots (AMR) depend on perception modules to interpret their surroundings, detect objects, estimate navigable free space, and avoid collisions. Previous research has shown that deep learning–based vision systems are effective for semantic navigation tasks such as object detection and scene understanding~\cite{technologies12090157}. More broadly, learning-based approaches, including deep learning and reinforcement learning, have been extensively studied for autonomous navigation~\cite{9764831}. However, their performance may degrade when deployed in environments that differ from the training distribution, particularly when human behavior, object arrangements, spatial layouts, and robot motion patterns vary. This limitation motivates the investigation of more general-purpose models for safety monitoring under realistic and uncertain navigation conditions. Vision–language models (VLMs) and related embodied multimodal models have recently gained significant attention in robotics because they can integrate visual observations with textual and state-based information~\cite{9764831}. PaLM-E~\cite{DBLP:journals/corr/abs-2303-03378} demonstrated that multimodal language models can combine visual and state representations for embodied reasoning tasks. These studies demonstrate the value of learning-based and multimodal models for robot perception and embodied reasoning, whereas our work investigates whether general-purpose VLMs can also support safety monitoring by identifying potentially unsafe situations and providing early warnings during robot navigation.

\subsection{VLMs for Robot Testing and Failure Monitoring}
\label{subsec:P2}
Recent software engineering research has begun exploring the use of VLMs for robot testing. For example, recent work~\cite{wu2026vlmTestingAMR} proposed VLM-based methods for testing industrial autonomous mobile robots by generating human behaviors that violate functional and safety requirements in simulation. Another work has introduced VLM-based test oracles~\cite{10.1145/3805760.3814910} to assess the correctness and quality of robot tasks from video observations. These efforts demonstrate that VLMs are promising tools for robot testing. However, they address different phases of the testing process, such as scenario generation and post hoc behavior assessment. In contrast, our work evaluates VLMs as safety-state predictors that classify navigation situations into safe, potentially unsafe, and unsafe categories to support early failure warning. 

Recent studies use classical approaches like deep learning models to detect and explain failures in robotic manipulation across different tasks and robots, while multitasking failure detector uses Vision-Language-Action (VLA) features to detect failures on unseen tasks and considers the trade-off between detection accuracy and warning time~\cite{technologies12090157,NEURIPS2025_392d0d05}. Predictive vision-language monitoring has also been used to estimate near-future execution risks and trigger fallback actions before collisions or workspace conflicts occur~\cite{10.3389/frobt.2026.1870024}, while failure prediction and correction VLA (FPC-VLA) uses a VLM-based supervisor to predict failures and generate corrective actions during manipulation~\cite{YANG2026131742}. 
VLMs have further been combined with digital twins for human--robot collaborative assembly, where the digital twin supports simulation and collision-risk mitigation~\cite{LIU2025100943}. Recent work has also investigated general-purpose foundation models as analytical components of digital twins. The study~\cite{10.3389/fpace.2026.1937868} benchmarks time-series foundation models across multiple predictive-maintenance tasks and shows that no single model performs best across all tasks in the zero-shot setting, highlighting the importance of selecting the foundation model according to the intended digital-twin task. Despite this progress, recent surveys still identify real-time monitoring and robustness as open safety challenges for VLM-enabled robots~\cite{10.1108/IR-02-2025-0074}. Building on these identified challenges, our work empirically evaluates the potential of VLMs for proactive robot-safety monitoring. We compare three general-purpose VLMs across three case studies, examining safety classification, the impact of sensor context and temporal history, prediction uncertainty and calibration, and the ability to detect unsafe situations early enough to provide actionable warnings, with the ultimate goal of enabling the use of VLMs in digital twins of robots.

\subsection{Prediction Reliability and Uncertainty}
Beyond detecting failures, safety-related robot monitoring also requires assessing whether a model's prediction can be trusted. The predicted label alone is insufficient; the model should also indicate whether its prediction is reliable. Reliability has also received attention: ViLU learns uncertainty estimates for distinguishing correct from incorrect VLM predictions~\cite{11444867}, while uncertainty-calibrated safety gating uses calibrated failure risk to decide when a VLA policy should continue, pause, or switch to a fallback controller~\cite{s26103140}. Classical approaches have shown that Maximum Softmax Probability (MSP) can serve as a simple baseline for identifying misclassified or out-of-distribution examples~\cite{DBLP:conf/iclr/HendrycksG17}. Calibration studies further show that modern neural networks can be poorly calibrated, meaning that their confidence scores may not accurately reflect their true likelihood of correctness~\cite{3305381.3305518}. We therefore use Expected Calibration Error (ECE) and Maximum Calibration Error (MCE) as calibration measures~\cite{3305381.3305518}. Recent work on VLM-based perception for autonomous underwater robotic software also highlights the importance of evaluating both performance and uncertainty when assessing whether VLMs are suitable for robotic perception tasks~\cite{yousaf2026vlmUnderwater,s26103140}. This is particularly important for VLM-based robot safety monitoring, where an overconfident wrong prediction may cause the system to miss an unsafe situation. Thus, dependable robot safety monitoring requires not only correct predictions but also reliable confidence, uncertainty, and calibration information. To this end, we evaluate classification performance alongside confidence, uncertainty, and calibration using MSP, Probability Confidence Score (PCS), entropy, DeepGini(DG), ECE, and MCE.

\subsection{Relation to Prior Work} 
\label{subsec:P3}
Prior work on learning-based robot perception and navigation~\cite{technologies12090157,9764831}, embodied multimodal reasoning~\cite{DBLP:journals/corr/abs-2303-03378}, and VLM-based robot testing and failure monitoring~\cite{wu2026vlmTestingAMR,10.1145/3805760.3814910,technologies12090157,NEURIPS2025_392d0d05} shows that these models can support robot perception, reasoning, testing, and failure detection. Our work addresses this gap by evaluating multiple VLMs using recent camera observations alone and using the same visual history together with robot-state information from odometry and LiDAR. Nevertheless, there remains limited empirical evidence on whether general-purpose VLMs can provide reliable early warnings before robot navigation failures occur.

\section{Context and Research Methodology}
\label{sec:methodology}
This section presents the context and methodology used to evaluate VLMs as perception and reasoning components of a robot digital twin. We first introduce the overall research context and then describe the four steps of the methodology, from case-study preparation to empirical analysis.

\subsection{Application Context}\label{sec:overview}

\begin{figure}[!t]
\centering
\includegraphics[width=1\textwidth]{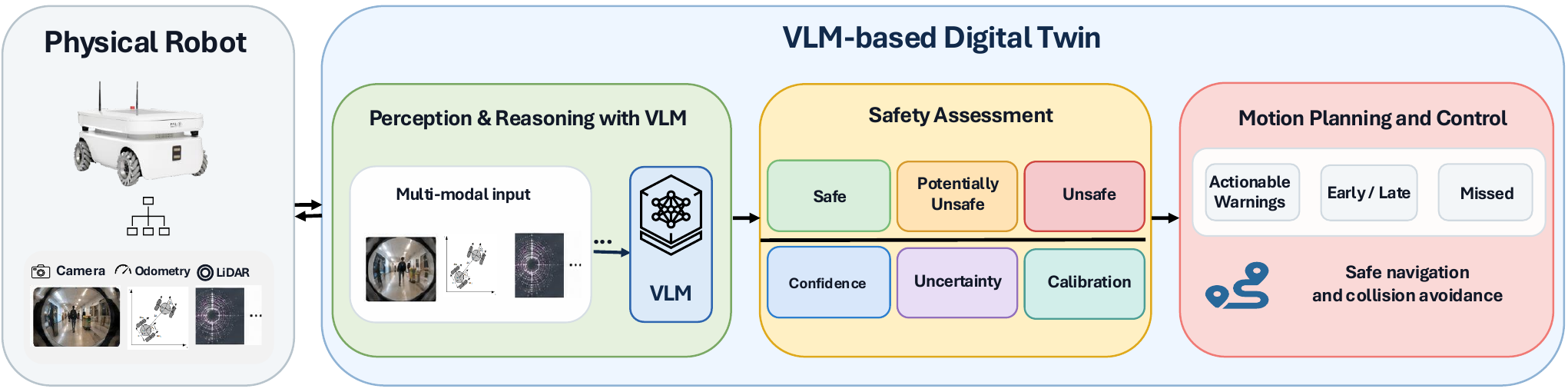}
\caption{Context of the VLM-based digital twin for robot-safety monitoring.}
\Description{}
\label{fig:dt_context}
\end{figure}

Figure~\ref{fig:dt_context} illustrates the application context of the VLM-based digital twin considered in this study.
The physical robot provides camera observations together with odometry and LiDAR information to the digital twin. The VLM processes these inputs to estimate the current safety state (\emph{safe}, \emph{potentially unsafe}, or \emph{unsafe}), together with confidence, uncertainty, and calibration information. These outputs are then used to assess warning timing and support motion-planning and safety decisions. In the intended operational setting, the control commands are sent back to the physical robot, such as changing velocity, stopping, or avoiding an obstacle. In the present study, however, we evaluate the perception, safety-assessment, and early-warning stages offline using recorded robot trajectories; therefore, the control commands are not executed and the VLM predictions do not modify the recorded robot motion. This subsection defines the VLM's intended role within the robot digital twin; Section~\ref {sec:experimental-design} presents the detailed experimental settings used to evaluate it.

\begin{figure}[!t]
\centering
\includegraphics[width=0.9\textwidth]{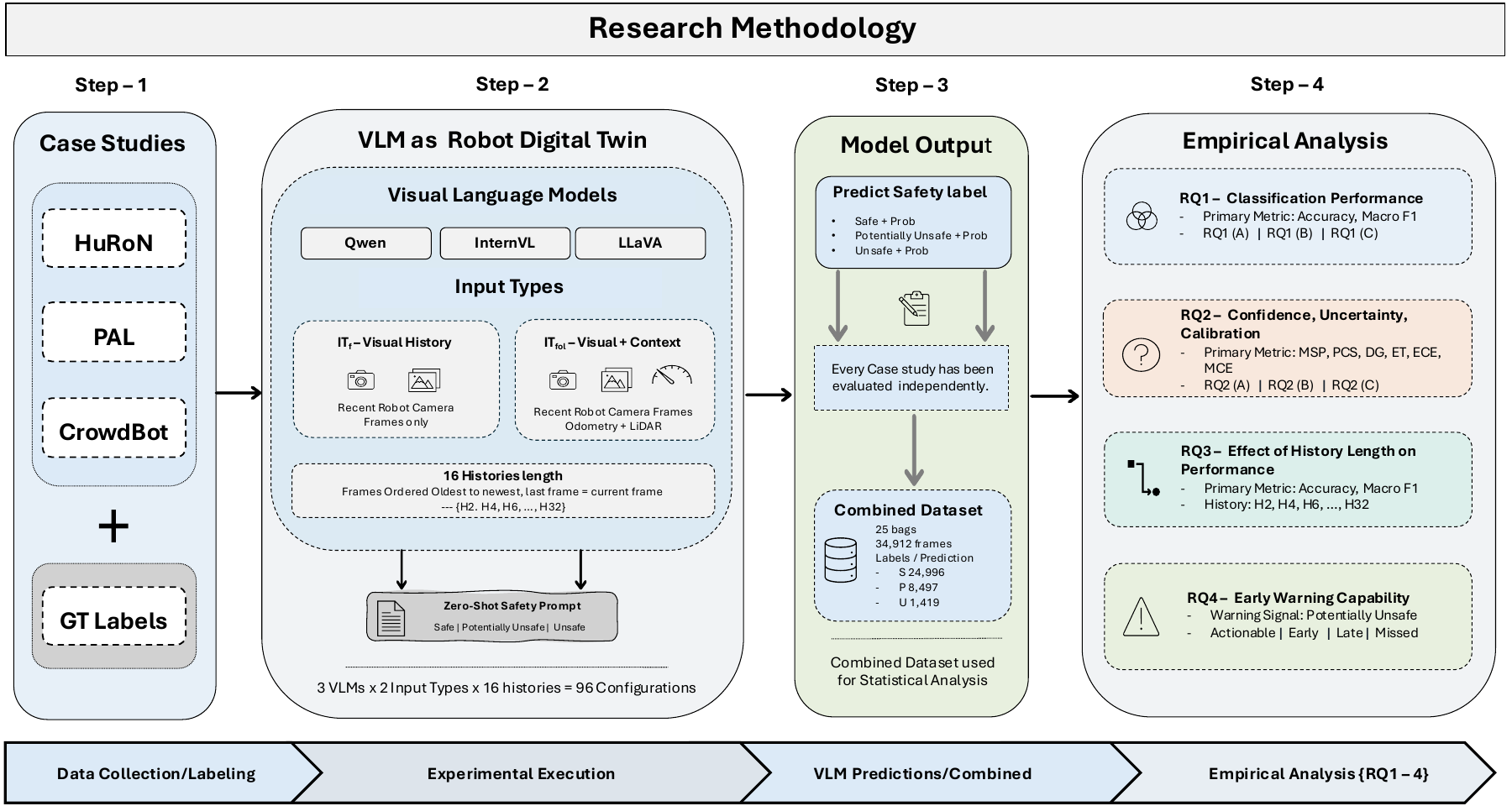}
\caption{Overview of the research methodology.}
\Description{}
\label{fig:RM}
\end{figure}

Figure~\ref{fig:RM} provides an overview of the research methodology organized into four main steps: (1) case study data collection and labeling, (2) experimental execution using VLMs as robot digital twins, (3) generation and aggregation of model outputs, and (4) empirical analysis addressing RQ1--RQ4.

\subsection{Step 1: Case Studies}
The empirical study considers three robot navigation case studies: HuRoN\footnote{HuRoN dataset: \url{https://rail.eecs.berkeley.edu/datasets/huron/}}, PAL Robotics/TIAGo \footnote{The PAL Robotics/TIAGo data were recorded by PAL Robotics in their own setting. This data is thus confidential and not provided online.}, and CrowdBot\footnote{CrowdBot dataset: \url{https://zenodo.org/records/17719415?preview_file=rosbags_0424_rds_defaced.zip}}. HuRoN captures indoor social-navigation scenarios in which a mobile robot moves through human-populated environments while avoiding pedestrians and nearby obstacles. The PAL case study uses a TIAGo mobile robot performing indoor navigation, where the robot must move safely while maintaining sufficient distance from surrounding obstacles. CrowdBot represents outdoor navigation in crowded pedestrian environments, where the robot moves among people and must avoid unsafe proximity and collisions. Together, these case studies provide 25 ROS bags comprising 34,912 labeled frames. Each frame is associated with one of three safety states: \emph{safe} (S), \emph{potentially unsafe} (P), or \emph{unsafe} (U). The case studies provide different robot-navigation scenarios and form the input data for the subsequent VLM experiments.

\subsection{Step 2: VLM as Robot Digital Twin}
\label{sec:RobotDT}

The second stage evaluates three VLMs---Qwen, InternVL, and LLaVA---as perception and reasoning components of a robot digital twin. We run the experiments offline on recorded robot trajectories; therefore, VLM predictions do not modify robot motion or the recorded observation sequence. For each labeled observation, we provide the VLM with a sequence of consecutive robot observations ending at the current observation. 

The history length is varied from H02 to H32 in increments of two frames, giving 16 history settings. Two input types are considered. \appFr{} uses only the recent history of robot camera frames, whereas \appOd{} combines the visual history with sensor information from odometry and LiDAR. These two input types were chosen to isolate the contribution of robot state information.

A common zero-shot safety prompt is used to classify the current navigation situation as \emph{safe}, \emph{potentially unsafe}, or \emph{unsafe}. Combining three VLMs, two input types, and 16 history lengths results in 96 experimental configurations. The detailed model selection, input-type, history construction, and prompt design are described in Section~\ref{sec:experimental-factors}.

\subsection{Step 3: Model Output and Data Aggregation}
For each experimental configuration, the VLM predicts a safety label together with the corresponding class probabilities. Each case study is evaluated independently before the predictions are combined for subsequent analysis. After inference, the predictions are combined with the corresponding ground-truth labels and relevant experimental information into a unified dataset. The aggregated dataset retains the same 34,912 observations and is subsequently used as the basis for the statistical analyses.

\subsection{Step 4: Empirical Analysis}
The final stage evaluates and analyzes the aggregated predictions to assess the overall behavior of the evaluated VLMs for robot safety monitoring. The analysis examines their predictive performance, the reliability of their outputs, the influence of temporal context, and their ability to provide timely warnings before unsafe situations occur. The detailed research questions and corresponding statistical analyses are introduced later in the experimental design.

\section{Experimental Design}
\label{sec:experimental-design}

Building on the methodology in Section~\ref{sec:methodology}, this section specifies the detailed experimental design used to evaluate the VLM configurations. The three case studies are prepared and evaluated independently using the same three safety classes: \emph{safe}, \emph{potentially unsafe}, and \emph{unsafe}. Their outputs are then combined for the main statistical analysis.

The remainder of this section is organized as follows. Section~\ref{sec:rqs} introduces the research questions; Section~\ref{sec:case-studies} describes the three case studies and their ground-truth preparation; Section~\ref{sec:combined-analysis} explains how the case-study outputs are combined; Section~\ref{sec:experimental-factors} presents the evaluated input types, models, and history lengths; Section~\ref{sec:warning-classification} defines the early-warning formulation; and Section~\ref{sec:evaluation} describes the evaluation measures and statistical analysis.

\subsection{Research Questions}
\label{sec:rqs}

We investigate four research questions:

\begin{itemize}
\item \textbf{RQ1 (Classification Performance):} How accurately do VLMs classify robot-navigation scenes as \emph{safe}, \emph{potentially unsafe}, or \emph{unsafe}?
\begin{itemize}
\item \textbf{RQ1(a):} How do \appFr{} and \appOd{} compare within each VLM?
\item \textbf{RQ1(b):} How do the VLMs compare within each input type?
\item \textbf{RQ1(c):} How do the VLMs compare across all evaluated configurations?
\end{itemize}
\item \textbf{RQ2 (Confidence and Uncertainty Reliability):} How well do VLM confidence, uncertainty, and calibration estimates reflect the reliability of robot-safety predictions?
\begin{itemize}
\item \textbf{RQ2(a):} Do confidence and uncertainty measures follow the expected relationship with prediction correctness?
\item \textbf{RQ2(b):} How do the VLMs compare in confidence, uncertainty, and calibration for each input type and history setting?
\item \textbf{RQ2(c):} How do the VLMs compare across all evaluated configurations for each reliability metric?
\end{itemize}
\item \textbf{RQ3 (History Length):} How does visual-history length affect VLM classification performance, and what history length provides reliable classification?
\item \textbf{RQ4 (Early-Warning Capability):} Can VLMs provide sufficiently early and actionable warnings before unsafe events occur?
\end{itemize}

These research questions examine complementary requirements for using VLMs in robot-safety monitoring. RQ1 establishes whether the models can correctly distinguish the three safety states and whether adding odometry and LiDAR information improves classification. This is important because a safety monitor must recognize not only common {\it safe} situations but also the less frequent {\it potentially unsafe} and {\it unsafe} situations. RQ2 examines whether the model's reported confidence and uncertainty are informative about prediction correctness. The expected relationship is that correct predictions should, on average, have higher confidence (MSP and PCS) and lower uncertainty (entropy and DG) than incorrect predictions, while calibration evaluates whether reported confidence agrees with observed correctness. This is important because an incorrect but highly confident safety prediction may be particularly problematic in a safety-monitoring setting. RQ3 investigates history length because additional temporal context may help the VLM understand how a situation is developing, but longer histories also increase computational cost and potential inference latency. Finally, RQ4 evaluates warning timing because detecting an unsafe situation is useful for intervention only if the warning is generated early enough for the robot to react, brake, or avoid the hazard.

\subsection{Ground-Truth Preparation}
\label{sec:case-studies}

The three case studies use different robots and provide different sensor information, including camera, LiDAR, and contact or proximity signals~\cite{rosBumperMsg}. We therefore generate the ground truth separately for each case study and then map all samples to the same three safety labels. This preserves the information available in each dataset while allowing the model outputs to be analyzed together.

\paragraph{HuRoN}
HuRoN contains 15 indoor-navigation bags selected from five environments (\texttt{bww1}, \texttt{bww2}, \texttt{bww8}, \texttt{cory1}, and \texttt{soda3}), with three bags from each environment. The selected sequences provide fisheye images together with robot-state and distance information and contain 16,862 frame-level samples. For HuRoN, the ground truth is generated from bumper/contact information and the closest pedestrian distance, with the fisheye videos used for manual verification. A frame is labeled \emph{unsafe} when the bumper/contact signal is active or the closest pedestrian distance is below 0.6\,m. A frame is labeled \emph{potentially unsafe} when no contact has occurred but the proximity signal is active, or when the closest pedestrian distance is between 0.6\,m and 1.2\,m. A frame is labeled \emph{safe} when there is no contact or proximity activation and the closest pedestrian distance is at least 1.2\,m. The safety distance margins are guided by the safety-distance formulation in~\cite{DBLP:journals/corr/abs-2508-07611}.

\paragraph{PAL Robotics/TIAGo.}
The PAL case study contains 3 bags and 9,249 samples. Ground truth labels are generated from the front laser scan. For each 2-second window, we compute the minimum valid front obstacle distance and compare it with thresholds selected with reference to the TIAGo OMNI Base footprint reported in the PAL Robotics handbook~\cite{palTiagoOmniHandbook}. A window is labeled \emph{unsafe} when the minimum distance is below 0.35\,m, \emph{potentially unsafe} when the distance is between 0.35\,m and 0.80\,m, and \emph{safe} when the distance is at least 0.80\,m. The unsafe threshold therefore represents a near-contact condition, while the potentially unsafe range represents a close obstacle for which avoidance is still possible.

\paragraph{CrowdBot.}
The CrowdBot case study contains 7 bags and 8,801 samples. Its ground truth is also based on the distance between the robot and nearby pedestrians or obstacles, but it uses the thresholds defined for the CrowdBot preprocessing pipeline. A sample is labeled \emph{unsafe} when the minimum distance is below 0.5\,m, \emph{potentially unsafe} when the distance is between 0.5\,m and 1.2\,m, and \emph{safe} when the distance is at least 1.2\,m. These case-specific labels are then mapped to the same three-class vocabulary used for HuRoN and PAL. 

Table~\ref{tab:dataset-summary} summarizes the final class distribution after dataset-specific cleaning and ground-truth preparation.
\begin{table}[tb]
\caption{Case study composition after dataset-specific preparation. The combined dataset is used for the main statistical analysis unless stated otherwise.}
\label{tab:dataset-summary}
\centering
\setlength{\tabcolsep}{10pt}
\begin{tabular}{lrrrrr}
\toprule
Case study & Bags & Frames & Safe & Potentially unsafe & Unsafe \\
\midrule
HuRoN & 15 & 16,862 & 15,273 & 1,378 & 211 \\
PAL & 3 & 9,249 & 5,518 & 2,769 & 962 \\
CrowdBot & 7 & 8,801 & 4,205 & 4,350 & 246 \\
\midrule
\textbf{Combined} & \textbf{25} & \textbf{34,912} &
\textbf{24,996} & \textbf{8,497} & \textbf{1,419} \\
\bottomrule
\end{tabular}
\end{table}
The class distribution differs considerably across the case studies: HuRoN is strongly dominated by \emph{safe} samples, PAL contains a relatively larger proportion of \emph{unsafe} samples, while CrowdBot contains similar numbers of \emph{safe} and \emph{potentially unsafe} samples. Overall, the \emph{unsafe} class represents only 4.06\% of the combined data, highlighting the class imbalance that is considered in the subsequent evaluation.

\subsection{Combined Analysis}
\label{sec:combined-analysis}

We first independently evaluated the three case studies because they use different robots, sensor information, and ground-truth definitions. However, all produce the same three safety labels and prediction probabilities. After this step, the cleaned outputs are combined while retaining the case study identifier and a unique bag identifier for each recorded bag. The combined dataset contains 34,912 samples from 25 bags: 15 from HuRoN, 3 from PAL, and 7 from CrowdBot.

Each model--input-type--history configuration is evaluated on every available bag. To reduce the influence of differing numbers of frames within recorded trajectories, the main comparisons are made on bag-level performance measures rather than directly on the pooled frames. The case study and bag identifiers are retained throughout the analysis so that the original data source remains traceable.

\subsection{Input Types, Models, History Lengths, and Prompt Design}
\label{sec:experimental-factors}

We evaluate two input types. \appFr{} uses visual history only, whereas \appOd{} uses the same visual history together with additional odometry and LiDAR information. These two input types are chosen to measure the added value of robot-state information while keeping the visual input unchanged. \appFr{} therefore serves as the visual-only baseline, whereas \appOd{} represents the sensor-enriched setting in which the VLM receives information about robot motion and nearby obstacles in addition to the camera history. 

For each input type, we evaluate Qwen2.5-VL, InternVL, and LLaVA-OneVision. These models were selected because they are open-weight, general-purpose VLMs from different model families, support multi-image inputs, and are available at a comparable 7--8B scale, enabling a controlled comparison under the same experimental design. A preliminary pilot study was conducted to determine a feasible temporal range. Increasing the number of input images substantially increased memory requirements and, in some cases, caused out-of-memory errors. H32 was the largest history length that could be evaluated consistently across the selected VLMs and configurations. We evaluate every second history length from H02 to H32 to provide regular coverage of the temporal range while keeping the number of experimental configurations computationally manageable. Each model is tested using 16 history lengths to systematically study how the temporal context affects the prediction of safety.
\[H \in\{2,4,6,\ldots,32\},\] denoted by \appFr{H} and \appOd{H}. The history includes the current frame. Thus, $H=2$ contains one preceding frame and the current frame, whereas $ H=32$ contains 31 preceding frames and the current frame. For each labeled observation $i$, the history ends at the current observation and contains no future information. The first observation is
\[ k=\max(1,i-H+1),\]
where $I_j$ denotes the camera image at observation $j$, giving the visual history
\[\mathcal{I}_{i}^{H}=\{I_k, \ldots, I_i\}.\]

At the beginning of a trajectory, when fewer than $H$ observations are available, the history starts from $I_1$ without padding. 

The complete experimental design therefore consists of $3$ models $\times$ $2$ input types $\times$ $16$ history lengths $=96$ configurations. Each combined configuration contains the same 25 bags and 34,912 samples. For \appFr{}, only $\mathcal{I}_{i}^{H}$ is provided. For \appOd{}, the same visual history is combined with the corresponding robot-state history $\mathcal{S}_{i}^{H}=\{S_k, \ldots, S_i\}$, where $s_j$ contains the odometry and LiDAR information associated with observation $j$.

All models are evaluated with a zero-shot prompt, without task-specific training or fine-tuning on the evaluated datasets. We use this setting to examine how well general-purpose VLMs can perform the safety-classification task using their pretrained knowledge.

Before selecting the final formulation, we experimented with different prompt versions that varied the description of image order, identification of the current frame, definitions of the three safety states, and presentation of odometry data. The prompt was derived from the three safety labels used in our study. We included only the information needed to interpret the image history and distinguish \emph{safe}, \emph{potentially unsafe}, and \emph{unsafe} situations. Numerical distance thresholds were not included because these thresholds differ across the case studies and are used to construct the ground truth. Including them in the prompt could therefore expose the labeling rules to the VLM. The same safety definitions were used for all evaluated VLMs, and the prompt was not tuned separately for individual models. The common prompt is
\begin{quote}
\small
``You are a robot safety monitor for robot navigation. The recent robot-camera frames are ordered from oldest to newest, with the last frame representing the current observation. Classify the current robot situation as \emph{safe}, \emph{potentially unsafe}, or \emph{unsafe}. A safe state has sufficient free space and requires no slowing, stopping, or avoidance. A potentially unsafe state is a warning state in which the scene is not yet unsafe but may become unsafe soon if the robot continues moving. An unsafe state represents contact, collision, extreme proximity to a person or obstacle, or an unavoidable immediate collision.''
\end{quote}

For \appOd{}, the common classification prompt is supplemented with information about the robot state derived from odometry and LiDAR, including linear and angular motion, obstacle distance, recent sensor context, and estimated stopping information. These values are used only as supporting context, and no future information is provided to the VLM.

\subsection{Early-Warning Formulation}
\label{sec:warning-classification}

For the early-warning analysis, a VLM prediction of \emph{potentially unsafe} is treated as a warning signal. For each unsafe event, $t_{\mathrm{warning}}$ denotes the time at which the VLM predicts \emph{potentially unsafe}, while $t_{\mathrm{unsafe}}$ denotes the start of the corresponding ground-truth unsafe event. The warning lead time \tLead is defined as
\begin{equation}
\tLead = t_{\mathrm{unsafe}} - t_{\mathrm{warning}}.
\label{eq:lead-time}
\end{equation}

The value of \tLead determines whether the warning is too early, actionable, too late, or missed.

In order to decide whether \tLead is sufficient for the robot to avoid the collision, we need to compute the minimum time required for the robot to react and stop; we do this by considering the braking dynamics. The braking time is derived from the basic constant-acceleration relation $v_f=v_i+at$~\cite{ling2016university}, where $v_i$ is the robot's velocity at the start of braking, $v_f$ is the final velocity, $a$ is the constant acceleration, and $t$ is the time required for the velocity to change from $v_i$ to $v_f$. For braking, $t$ corresponds to the braking time $T_{\mathrm{brake}}$. Assuming constant braking deceleration and setting $v_f=0$, the braking time is given by
\begin{equation}
0 = v_i + a_{\mathrm{brake}}T_{\mathrm{brake}},
\end{equation}
and therefore 
\begin{equation}
T_{\mathrm{brake}} = -\frac{v_i}{a_{\mathrm{brake}}},
\label{eq:braking-time}
\end{equation}
where $v_i$ is the robot velocity at the start of braking, obtained from the recorded odometry data, and $a_{\mathrm{brake}}<0$ is the assumed constant braking acceleration.

Accounting for the assumed reaction time and an additional safety margin, the minimum required warning lead time is defined as
\begin{equation}
T_{\mathrm{required}} = T_{\mathrm{reaction}} + T_{\mathrm{brake}} + T_{\mathrm{margin}},
\label{eq:required-warning-time}
\end{equation}
where $T_{\mathrm{reaction}}$ is the assumed reaction time, $T_{\mathrm{brake}}$ is the braking time defined in Eq.~\eqref{eq:braking-time}, and $T_{\mathrm{margin}}$ is an additional safety margin.

To prevent warnings generated arbitrarily far in advance from being counted as successful, we define the earliest accepted warning lead time as
\begin{equation}
T_{\mathrm{earliest}} = T_{\mathrm{required}}+A,
\label{eq:earliest-warning-time}
\end{equation}
where the allowed interval is varied as $A\in\{1,$ $2,$ $3,$ $4,$ $5\}$\,s.

The warning outcome is then defined as
\begin{equation}
\mathrm{Outcome}= \begin{cases}
\mathrm{Too\ early}, & \tLead > \tEarliest,\\
\mathrm{Actionable}, & \tRequired \leq \tLead \leq \tEarliest,\\
\mathrm{Too\ late}, & 0<\tLead< \tRequired,\\
\mathrm{Missed}, & \tLead\leq0\ \text{or no warning is generated}.
\end{cases}
\label{eq:warning-outcome}
\end{equation}

An actionable warning therefore occurs early enough to allow stopping but within the accepted warning interval. A too-early warning is generated before the accepted interval, while a too-late warning is generated before the unsafe event but without enough time for the robot to react and stop. A missed warning means that no warning is generated before the unsafe event, or that it is generated only at or after the unsafe event has already started. Warning coverage is the proportion of unsafe events for which any warning is produced before the event, regardless of whether it is too early, actionable, or too late.

\subsection{Evaluation Measures and Statistical Analysis}
\label{sec:evaluation}

All main statistical comparisons use the bag as the analysis unit. The same bags are evaluated across models, input types, and history lengths. We use a significance level of $\alpha=0.05$ and apply Holm correction within each family of multiple comparisons. Rank-biserial correlation ($r_{\mathrm{rb}}$) is reported as effect size for paired Wilcoxon tests, and Kendall's $W$ is reported as effect size for Friedman tests.

\paragraph{\textbf{RQ1: Classification performance.}}
We report both Accuracy and Macro-F1 as the main overall metrics, together with per-class precision, recall, and F1. Accuracy and Macro-F1 are defined as
\begin{equation}
\mathrm{Accuracy}=\frac{\sum_{c\in\mathcal{C}}TP_c}{N}, \qquad \mathrm{MacroF1}=\frac{1}{|\mathcal{C}|}\sum_{c\in\mathcal{C}}F1_c.
\label{eq:classification-metrics}
\end{equation}
Here, $\mathcal{C}=\{\emph{safe},\emph{potentially\ unsafe},\emph{unsafe}\}$ is the set of safety classes, $TP_c$ is the number of correctly predicted samples for class $c$, $N$ is the total number of samples, $F1_c$ is the F1-score for class $c$, and $|\mathcal{C}|=3$ is the number of classes. Macro-F1 is important because the combined dataset is class-imbalanced. For each model and history, \appFr{} and \appOd{} are compared with paired Wilcoxon signed-rank tests over the 25 bags. Model differences are evaluated using Friedman tests followed, when significant, by Holm-adjusted paired Wilcoxon tests with rank-biserial correlation ($r_{\mathrm{rb}}$) reported as the corresponding effect size.

\paragraph{\textbf{RQ2: Confidence, uncertainty, and calibration.}}
For the predicted probability vector $\mathbf{p}$, we compute Maximum Softmax Probability (MSP), Probability Confidence Score (PCS), entropy, and DeepGini (DG). MSP is the highest predicted class probability, while PCS measures the gap between the two most probable classes. Entropy measures how spread the probability distribution is across all classes, and DG measures the overall uncertainty of the class-probability distribution:
\begin{equation}
\mathrm{MSP}=\max_c p_c,\quad \mathrm{PCS}=p_{(1)}-p_{(2)},\quad H(\mathbf{p})=-\sum_c p_c\log p_c,\quad \mathrm{DG}(\mathbf{p})=1-\sum_c p_c^2.
\label{eq:uncertainty-metrics}
\end{equation}
Here, $p_c$ is the predicted probability of class $c$, and $p_{(1)}$ and $p_{(2)}$ denote the largest and second-largest class probabilities, respectively. The maximum in MSP and the summations in entropy and DG are taken over the three safety classes. $H(\mathbf{p})$ denotes the entropy of the predicted probability distribution, while $\mathrm{DG}(\mathbf{p})$ denotes its DG uncertainty. While MSP, PCS, entropy, and DG characterize the confidence or uncertainty of individual predictions, calibration evaluates whether the reported confidence is consistent with the model's observed correctness. Calibration is evaluated using Expected Calibration Error (ECE) and Maximum Calibration Error (MCE). Let the predictions be divided into $M$ confidence bins, where $B_m$ denotes the set of predictions in bin $m$, and $N$ is the total number of predictions. For each bin, $\mathrm{acc}(B_m)$ is the observed accuracy and $\mathrm{conf}(B_m)$ is the average predicted confidence. We compute
\begin{equation}
\mathrm{ECE}
=
\sum_{m=1}^{M}
\frac{|B_m|}{N}
\left|
\mathrm{acc}(B_m)-\mathrm{conf}(B_m)
\right|,
\qquad
\mathrm{MCE}
=
\max_{m}
\left|
\mathrm{acc}(B_m)-\mathrm{conf}(B_m)
\right|.
\label{eq:calibration-metrics}
\end{equation}
Higher MSP and PCS mean that the VLM is more confident in its predicted safety state, whereas higher entropy and DG mean that it is more uncertain between the \emph{safe}, \emph{potentially unsafe}, and \emph{unsafe} classes. In our setting, high confidence is useful when the prediction is correct, but an overconfident prediction of \emph{safe} for an actually unsafe situation is particularly problematic because it may lead to a missed warning. Higher uncertainty indicates that the model finds the current situation ambiguous, but it does not by itself mean that the situation is unsafe.

ECE measures the average difference between predicted confidence and observed correctness, while MCE captures the largest such difference across the confidence bins. Lower ECE and MCE therefore indicate better calibration and make the reported confidence more trustworthy when interpreting a safety prediction. 

For RQ2 (a), we use point-biserial correlation because prediction correctness is binary (correct or incorrect), whereas MSP, PCS, entropy, and DG are continuous measures. This allows us to assess whether correct predictions tend to have higher confidence and lower uncertainty than incorrect predictions. For RQ2(b) and RQ2(c), model comparisons use bag-level measures and the same paired non-parametric testing framework as RQ1. 
\paragraph{\textbf{RQ3: History-length effect.}}
For each model--input type combination, the 16 histories are compared using a Friedman test over the 25 bag-level Accuracy and Macro-F1. The six Friedman tests within each metric are Holm-corrected. When a Friedman test remains significant, the 120 possible history pairs are compared with paired Wilcoxon signed-rank tests and Holm correction. Histories are then summarized by their significant wins and losses.

\paragraph{\textbf{RQ4: Early-warning capability.}}
RQ4 evaluates whether VLM predictions of \emph{potentially unsafe} provide sufficiently early warnings before ground-truth unsafe events. To avoid selecting configurations based on their early-warning outcomes, the evaluated histories are preselected using the RQ3 classification analysis. For each selected configuration and unsafe event, the timing of the \emph{potentially unsafe} prediction is evaluated according to the early-warning formulation in Section~\ref{sec:warning-classification}. Each event is classified as actionable, too early, too late, or missed, and warning coverage is also computed.

\section{Results}\label{sec:results}
This section presents the empirical results for classification performance, prediction reliability, the effect of history length, and early-warning capability, following RQ1--RQ4.
\subsection{RQ1: Classification Performance}\label{sec:rq1}

In this RQ, we are interested in the classification performance of the evaluated VLMs across the three safety states: \emph{safe}, \emph{potentially unsafe}, and \emph{unsafe}.

\subsubsection{Descriptive Analysis}
\label{sec:rq1_descriptive}

Table~\ref{tab:rq1_descriptive_main_percent} summarizes pooled classification performance across histories.
\begin{table}[!tb]
\caption{RQ1 -- Pooled classification performance across the combined evaluation data by model, input type, and history. All values are percentages.}
\label{tab:rq1_descriptive_main_percent}
\centering
\small
\setlength{\tabcolsep}{4pt}
\begin{tabular}{lrrrrrrrrrrrr}
\toprule
& \multicolumn{4}{c}{Qwen} & \multicolumn{4}{c}{InternVL} & \multicolumn{4}{c}{LLaVA} \\
\cmidrule(lr){2-5}\cmidrule(lr){6-9}\cmidrule(lr){10-13}
& \multicolumn{2}{c}{\appFr{}} & \multicolumn{2}{c}{\appOd{}} & \multicolumn{2}{c}{\appFr{}} & \multicolumn{2}{c}{\appOd{}} & \multicolumn{2}{c}{\appFr{}} & \multicolumn{2}{c}{\appOd{}} \\
\cmidrule(lr){2-3}\cmidrule(lr){4-5}\cmidrule(lr){6-7}\cmidrule(lr){8-9}\cmidrule(lr){10-11}\cmidrule(lr){12-13}
History & Acc. & MF1 & Acc. & MF1 & Acc. & MF1 & Acc. & MF1 & Acc. & MF1 & Acc. & MF1 \\
\midrule
H02 & \textbf{71.98} & \textbf{43.31} & 68.64 & 40.68 & 66.11 & 41.72 & 59.35 & 38.00 & 61.13 & 38.81 & 60.18 & 36.02 \\
H04 & \textbf{71.70} & 40.92 & 65.71 & \textbf{41.61} & 65.72 & 41.05 & 58.84 & 38.15 & 65.23 & 38.42 & 63.92 & 36.69 \\
H06 & \textbf{72.33} & 42.04 & 70.29 & \textbf{43.66} & 54.29 & 35.91 & 50.29 & 33.79 & 65.92 & 37.15 & 64.64 & 36.82 \\
H08 & \textbf{72.05} & 41.02 & 68.95 & \textbf{42.76} & 54.58 & 35.96 & 50.13 & 33.73 & 66.01 & 35.63 & 64.83 & 35.79 \\
H10 & \textbf{72.51} & 42.38 & 70.49 & \textbf{43.54} & 50.15 & 34.02 & 52.91 & 34.88 & 65.84 & 35.85 & 64.06 & 37.03 \\
H12 & \textbf{72.26} & 41.86 & 70.49 & \textbf{43.60} & 51.04 & 34.19 & 52.69 & 34.66 & 64.74 & 36.58 & 63.46 & 36.26 \\
H14 & \textbf{72.40} & 42.38 & 71.02 & \textbf{43.41} & 63.59 & 40.12 & 61.01 & 39.17 & 64.90 & 36.67 & 63.41 & 36.72 \\
H16 & \textbf{72.33} & 42.30 & 70.49 & \textbf{43.05} & 63.49 & 40.06 & 59.45 & 38.47 & 65.98 & 36.05 & 64.53 & 35.09 \\
H18 & \textbf{72.86} & 41.70 & 71.26 & \textbf{43.27} & 48.86 & 32.94 & 53.25 & 34.97 & 66.27 & 36.17 & 63.88 & 36.34 \\
H20 & \textbf{72.87} & 41.57 & 69.93 & \textbf{42.74} & 51.98 & 34.68 & 53.53 & 35.35 & 66.03 & 36.45 & 63.99 & 34.76 \\
H22 & \textbf{73.07} & 41.66 & 70.62 & \textbf{43.27} & 50.33 & 33.92 & 55.29 & 36.05 & 65.95 & 35.86 & 63.55 & 36.63 \\
H24 & \textbf{73.03} & 41.67 & 71.05 & \textbf{43.23} & 50.11 & 33.50 & 54.71 & 35.84 & 66.26 & 35.38 & 63.43 & 34.96 \\
H26 & \textbf{73.24} & 41.59 & 71.69 & \textbf{44.16} & 50.68 & 33.91 & 52.68 & 35.01 & 66.28 & 34.94 & 63.58 & 34.38 \\
H28 & \textbf{73.37} & 41.76 & 71.52 & \textbf{44.24} & 53.02 & 35.09 & 52.05 & 34.77 & 66.54 & 35.02 & 64.11 & 33.58 \\
H30 & \textbf{73.19} & 40.74 & 72.22 & \textbf{43.91} & 53.10 & 35.35 & 56.87 & 37.72 & 66.77 & 35.21 & 64.31 & 33.92 \\
H32 & \textbf{73.19} & 40.88 & 72.78 & \textbf{44.87} & 54.16 & 35.77 & 55.45 & 37.03 & 66.85 & 34.99 & 63.98 & 34.49 \\
\bottomrule
\end{tabular}
\par\smallskip
\begin{minipage}{\linewidth}
\footnotesize
\textit{Note.} Acc. = accuracy; MF1 = Macro-F1. Bold identifies the highest value for each metric within each history across the six model--input type combinations; it does not indicate statistical significance.
\end{minipage}
\end{table}
Qwen with \appFr{} achieves the highest accuracy at every history (71.70--73.37\%), whereas Qwen with \appOd{} achieves the highest Macro-F1 at 15 out of 16 histories and reaches 44.87\% at H32. The per-class results, reported in the supplementary material~\cite{farooq2026vlmrepo}, show that this overall performance is uneven across safety states. At H32, Qwen with \appFr{} attains 91.71\% recall for safe scenes but 30.94\% for potentially unsafe scenes and 0\% for unsafe scenes. At the same history, \appOd{} increases potentially unsafe recall to 50.43\% and F1 to 50.10\%, producing a higher Macro-F1 than \appFr{} (44.87\% versus 40.88\%). Across all model--input type--history configurations, unsafe recall never exceeds 1.06\%. Thus, accuracy should be interpreted together with Macro-F1 and per-class recall.

\subsubsection{RQ1(a): Input Type Comparison}
\label{sec:rq1a_Input Type_comparison}

Table~\ref{tab:rq1a_approach_comparison} summarizes within-model \appFr{}--\appOd{} comparisons using paired Wilcoxon signed-rank tests with Holm correction.
\begin{table}[!tb]
\caption{RQ1(a) -- Significant within-model differences between \appFr{} and \appOd{} across history lengths.}
\label{tab:rq1a_approach_comparison}
\centering
\small
\begin{tabular}{lcccccc}
\toprule
& \multicolumn{2}{c}{Qwen}
& \multicolumn{2}{c}{InternVL}
& \multicolumn{2}{c}{LLaVA} \\
\cmidrule(lr){2-3}\cmidrule(lr){4-5}\cmidrule(lr){6-7}
History & Accuracy & Macro-F1 & Accuracy & Macro-F1 & Accuracy & Macro-F1 \\
\midrule
H02 & \textemdash & \textemdash & \appFr{} & \textemdash & \textemdash & \textemdash \\
H04 & \appFr{} & \textemdash & \appFr{} & \textemdash & \textemdash & \textemdash \\
H06 & \textemdash & \textemdash & \textemdash & \textemdash & \textemdash & \textemdash \\
H08 & \textemdash & \textemdash & \textemdash & \textemdash & \textemdash & \textemdash \\
H10 & \textemdash & \textemdash & \textemdash & \appOd{} & \textemdash & \textemdash \\
H12 & \textemdash & \textemdash & \textemdash & \textemdash & \textemdash & \textemdash \\
H14 & \textemdash & \textemdash & \textemdash & \textemdash & \textemdash & \textemdash \\
H16 & \textemdash & \textemdash & \appFr{} & \textemdash & \textemdash & \textemdash \\
H18 & \textemdash & \textemdash & \appOd{} & \appOd{} & \textemdash & \textemdash \\
H20 & \textemdash & \textemdash & \textemdash & \textemdash & \textemdash & \textemdash \\
H22 & \textemdash & \textemdash & \appOd{} & \appOd{} & \textemdash & \textemdash \\
H24 & \textemdash & \textemdash & \appOd{} & \appOd{} & \appFr{} & \textemdash \\
H26 & \textemdash & \textemdash & \textemdash & \textemdash & \appFr{} & \textemdash \\
H28 & \textemdash & \textemdash & \textemdash & \textemdash & \textemdash & \textemdash \\
H30 & \textemdash & \textemdash & \textemdash & \textemdash & \appFr{} & \textemdash \\
H32 & \textemdash & \textemdash & \textemdash & \textemdash & \appFr{} & \textemdash \\
\bottomrule
\end{tabular}
\par\smallskip
\begin{minipage}{\linewidth}
\footnotesize
\textit{Note.} An entry identifies the input type with significantly higher performance according to the paired Wilcoxon signed-rank test after Holm correction ($p_{\mathrm{Holm}}<0.05$). \textemdash{} indicates no statistically significant difference.
\end{minipage}
\end{table}
Significant differences occur in 15 of the 96 model--history--metric comparisons. For Qwen, \appFr{} yields higher accuracy only at H04 ($p_{\mathrm{Holm}}=0.032$, $r_{\mathrm{rb}}=0.680$), with no significant Macro-F1 differences. For InternVL, \appFr{} yields higher accuracy at H02, H04, and H16, whereas \appOd{} yields higher accuracy at H18, H22, and H24; \appOd{} also improves Macro-F1 at H10, H18, H22, and H24, with the largest effect at H24 ($p_{\mathrm{Holm}}=0.001$, $r_{\mathrm{rb}}=-0.840$). For LLaVA, \appFr{} yields higher accuracy at H24, H26, H30, and H32, while Macro-F1 differences are non-significant throughout. Positive and negative $r_{\mathrm{rb}}$ values favor \appFr{} and \appOd{}, respectively. The complete results for all 96 comparisons, including the Wilcoxon statistic, raw and Holm-adjusted $p$-values, median differences, and rank-biserial effect sizes, are provided in the supplementary material~\cite{farooq2026vlmrepo}.

\subsubsection{RQ1(b): Model Comparison}\label{sec:rq1b_model_comparison}

Across the 25 paired bags, Holm-adjusted Friedman tests identify model differences in accuracy at all 16 histories for both input types, in Macro-F1 at 13 of 16 \appFr{} histories, and in Macro-F1 at all 16 \appOd{} histories. Table~\ref{tab:rq1b_model_comparison} summarizes the post-hoc results by showing, for each input type, metric, and history, whether one model significantly outperforms both other models. A model is reported only when the Friedman test shows a significant difference and both Holm-adjusted pairwise Wilcoxon comparisons are significant; a dash indicates that no model satisfies both conditions. Post-hoc paired Wilcoxon tests identify Qwen as the only model that significantly outperforms both competitors in accuracy: at all 16 \appFr{} histories and 15 out of 16 \appOd{} histories.
\begin{table}[!tb]
\caption{RQ1(b) -- Models that significantly outperform both competitors within each input type and history.}
\label{tab:rq1b_model_comparison}
\centering
\footnotesize
\setlength{\tabcolsep}{1pt}
\begin{tabular*}{\linewidth}{@{\extracolsep{\fill}}ll*{16}{c}@{}}
\toprule
& & \multicolumn{16}{c}{History} \\
\cmidrule(lr){3-18}
Approach & Metric & H02 & H04 & H06 & H08 & H10 & H12 & H14 & H16 & H18 & H20 & H22 & H24 & H26 & H28 & H30 & H32 \\
\midrule
\appFr{} & Accuracy & \textbf{Q} & \textbf{Q} & \textbf{Q} & \textbf{Q} & \textbf{Q} & \textbf{Q} & \textbf{Q} & \textbf{Q} & \textbf{Q} & \textbf{Q} & \textbf{Q} & \textbf{Q} & \textbf{Q} & \textbf{Q} & \textbf{Q} & \textbf{Q} \\
& Macro-F1 & \textemdash & \textemdash & \textemdash & \textemdash & \textemdash & \textemdash & \textemdash & \textemdash & \textemdash & \textemdash & \textemdash & \textemdash & \textemdash & \textemdash & \textemdash & \textemdash \\
\addlinespace
\appOd{} & Accuracy & \textbf{Q} & \textemdash & \textbf{Q} & \textbf{Q} & \textbf{Q} & \textbf{Q} & \textbf{Q} & \textbf{Q} & \textbf{Q} & \textbf{Q} & \textbf{Q} & \textbf{Q} & \textbf{Q} & \textbf{Q} & \textbf{Q} & \textbf{Q} \\
& Macro-F1 & \textbf{Q} & \textemdash & \textbf{Q} & \textbf{Q} & \textbf{Q} & \textbf{Q} & \textbf{Q} & \textemdash & \textemdash & \textbf{Q} & \textemdash & \textemdash & \textbf{Q} & \textbf{Q} & \textbf{Q} & \textbf{Q} \\
\bottomrule
\end{tabular*}
{
\begin{flushleft}
\footnotesize
\textit{Note.} Q denotes Qwen. A model is shown only when the Friedman test remains significant after Holm correction and that model significantly outperforms both other models in the Holm-adjusted pairwise Wilcoxon tests ($p_{\mathrm{Holm}}<0.05$). \textemdash{} indicates that no model satisfies both conditions; individual model pairs may nevertheless differ significantly.
\end{flushleft}
}
\end{table}
For \appFr{} Macro-F1, no model beats both competitors, although Qwen and LLaVA each outperform InternVL at 12 histories. Under \appOd{}, Qwen beats both competitors in Macro-F1 at 11 histories; at H32, its advantages over LLaVA and InternVL are $r_{\mathrm{rb}}=0.557$ and $0.760$, respectively. Overall, Qwen shows the most consistent accuracy advantage, while its Macro-F1 advantage depends more strongly on input type and history. Full Friedman and post-hoc results are provided in the supplementary material~\cite{farooq2026vlmrepo}.

\subsubsection{RQ1(c): Overall Model Comparison}\label{sec:rq1c_overall_model_comparison}

Across 800 matched bag--input type--history configurations, Friedman tests identify model differences for Macro-F1 ($\chi_F^2(2)=371.19$, $p=2.49\times10^{-81}$, $W=0.232$) and accuracy ($\chi_F^2(2)=677.34$, $p=8.29\times10^{-148}$, $W=0.423$). All six Holm-adjusted pairwise Wilcoxon comparisons are significant, giving the same aggregate ordering for both metrics: $\mathrm{Qwen}$ $>\mathrm{LLaVA}$ $>$ $\mathrm{InternVL}$ (Table~\ref{tab:rq1c_overall_model_comparison}).
\begin{table}[!tb]
\caption{RQ1(c) -- Aggregate pairwise model comparisons across 800 matched experimental configurations.}
\label{tab:rq1c_overall_model_comparison}
\centering
\small
\begin{tabular}{lllllll}
\toprule
Metric & Model 1 & Model 2 & $r_{\mathrm{rb}}$ & Effect size & Better model & $p_{\mathrm{Holm}}$ \\
\midrule
Macro-F1
& Qwen  & LLaVA    & 0.284 & Medium & Qwen  & $3.63\times10^{-12}$ \\
& Qwen  & InternVL & 0.686 & Large  & Qwen  & $6.02\times10^{-63}$ \\
& LLaVA & InternVL & 0.578 & Large  & LLaVA & $3.02\times10^{-45}$ \\
\addlinespace
Accuracy
& Qwen  & LLaVA    & 0.804 & Large & Qwen  & $5.13\times10^{-86}$ \\
& Qwen  & InternVL & 0.859 & Large & Qwen  & $1.04\times10^{-97}$ \\
& LLaVA & InternVL & 0.591 & Large & LLaVA & $2.50\times10^{-47}$ \\
\bottomrule
\end{tabular}

\par\smallskip
{
\footnotesize
\begin{flushleft}
\textit{Note.} $r_{\mathrm{rb}}$ is the matched-pairs rank-biserial correlation for Model~1 minus Model~2; positive values favor Model~1. Effect-size magnitudes are interpreted using $|r_{\mathrm{rb}}|$ as small ($\geq0.11$), medium ($\geq0.28$), and large ($\geq0.43$). All pairwise comparisons are significant after Holm correction ($p_{\mathrm{Holm}}<0.05$). The 800 matched configurations comprise repeated evaluations of the same 25 bags across two approaches and 16 histories.
\end{flushleft}
}
\end{table}

Qwen's advantage over LLaVA is larger for accuracy ($r_{\mathrm{rb}}=0.804$) than for Macro-F1 ($r_{\mathrm{rb}}=0.284$). Because the 800 configuration values correspond to the same 25 bags evaluated under 2 input types and 16 histories, this analysis represents an aggregate comparison across repeated experimental configurations rather than 800 independent units. Complete test statistics are provided in the supplementary material~\cite{farooq2026vlmrepo}.

Taken together, RQ1 identifies Qwen as the strongest overall classifier, particularly in accuracy, while the relative benefit of \appFr{} and \appOd{} varies by model and metric. The main limitation is the recognition of unsafe states: unsafe recall is at most 1.06\% across all evaluated configurations. Accordingly, overall accuracy must be interpreted together with Macro-F1 and per-class recall.

\answerRQ{
\textbf{Answer to RQ1.}
Qwen provides the strongest overall classification performance, with the most consistent advantage in Accuracy and an aggregate ordering of $\mathrm{Qwen}$ $>$ $\mathrm{LLaVA}$ $>$ $\mathrm{InternVL}$. The benefit of adding odometry and LiDAR is model- and metric-dependent. For Qwen, \appOd{} provides stronger class-balanced performance in many settings, while \appFr{} often retains higher Accuracy. 
}

\subsection{RQ2: Confidence and Uncertainty Reliability}
\label{sec:rq2}

In this RQ, we are interested in how well VLM confidence, uncertainty, and calibration reflect prediction reliability.

\subsubsection{RQ2(a): Correctness--Confidence/Uncertainty Association}
\label{sec:rq2a_correctness_association}

RQ2(a) tests whether confidence and uncertainty estimates track prediction correctness. Correctness is coded as 1 for a correct prediction and 0 otherwise, and the point-biserial correlation ($r_{\mathrm{pb}}$) is computed between correctness and each frame-level estimate across the 25 evaluation bags. The expected reliability direction is positive for MSP/PCS and negative for DG (DG)/entropy.

Across both input types, Qwen follows the expected reliability direction at every history. Under \appFr{}, MSP ranges from 0.294 to 0.352, PCS from 0.290 to 0.347, DG from $-0.359$ to $-0.304$, and entropy from $-0.363$ to $-0.309$ (Table~\ref{tab:rq2a_appfr}). LLaVA also follows the expected direction for every available coefficient, although the associations are generally weaker than those of Qwen. InternVL shows a less stable pattern: under \appFr{}, all available coefficients are in the opposite direction, with absolute correlations reaching approximately $0.47$--$0.52$ at H26--H32.

Under \appOd{}, Qwen and LLaVA again follow the expected reliability direction at every history. InternVL follows the expected direction at H02--H04, shows the opposite direction at H06--H24, and returns to the expected direction at H26--H32 (Table~\ref{tab:rq2a_appod}).

\begin{table}[!tb]
\caption{RQ2(a) -- Frame-level point-biserial correlations between prediction correctness and confidence/uncertainty estimates. Bold values indicate the expected reliability direction: positive for confidence measures and negative for uncertainty measures.}
\label{tab:rq2a}
\centering
\scriptsize
\setlength{\tabcolsep}{2.2pt}

\begin{subtable}[t]{\linewidth}
\centering
\caption{\appFr{}}
\label{tab:rq2a_appfr}

\begin{tabular*}{\linewidth}{@{\extracolsep{\fill}}l*{12}{r}@{}}
\toprule
& \multicolumn{4}{c}{Qwen} & \multicolumn{4}{c}{InternVL} & \multicolumn{4}{c}{LLaVA} \\
\cmidrule(lr){2-5}\cmidrule(lr){6-9}\cmidrule(lr){10-13}
History & MSP & PCS & DG & Ent. & MSP & PCS & DG & Ent. & MSP & PCS & DG & Ent. \\
\midrule
H02 & \textbf{0.294} & \textbf{0.290} & \textbf{-0.304} & \textbf{-0.309} & \textemdash & \textemdash & \textemdash & \textemdash & \textbf{0.055} & \textbf{0.047} & \textemdash & \textbf{-0.063} \\
H04 & \textbf{0.330} & \textbf{0.326} & \textbf{-0.346} & \textbf{-0.353} & -0.032 & -0.032 & 0.033 & 0.035 & \textbf{0.247} & \textbf{0.240} & \textbf{-0.207} & \textbf{-0.259} \\
H06 & \textbf{0.309} & \textbf{0.304} & \textbf{-0.322} & \textbf{-0.328} & -0.056 & -0.056 & 0.058 & 0.063 & \textbf{0.268} & \textbf{0.262} & \textbf{-0.230} & \textbf{-0.279} \\
H08 & \textbf{0.330} & \textbf{0.326} & \textbf{-0.345} & \textbf{-0.354} & -0.082 & -0.082 & 0.083 & 0.087 & \textbf{0.275} & \textbf{0.268} & \textbf{-0.248} & \textbf{-0.301} \\
H10 & \textbf{0.331} & \textbf{0.327} & \textbf{-0.341} & \textbf{-0.345} & -0.089 & -0.089 & 0.090 & 0.090 & \textbf{0.278} & \textbf{0.270} & \textbf{-0.262} & \textbf{-0.316} \\
H12 & \textbf{0.352} & \textbf{0.347} & \textbf{-0.359} & \textbf{-0.363} & -0.020 & -0.020 & 0.020 & 0.022 & \textbf{0.249} & \textbf{0.241} & \textbf{-0.220} & \textbf{-0.287} \\
H14 & \textbf{0.344} & \textbf{0.334} & \textbf{-0.354} & \textbf{-0.359} & \textemdash & \textemdash & \textemdash & 0.015 & \textbf{0.255} & \textbf{0.247} & \textbf{-0.229} & \textbf{-0.292} \\
H16 & \textbf{0.346} & \textbf{0.335} & \textbf{-0.355} & \textbf{-0.361} & \textemdash & \textemdash & \textemdash & \textemdash & \textbf{0.274} & \textbf{0.266} & \textbf{-0.260} & \textbf{-0.320} \\
H18 & \textbf{0.322} & \textbf{0.312} & \textbf{-0.332} & \textbf{-0.337} & -0.040 & -0.040 & 0.041 & 0.043 & \textbf{0.268} & \textbf{0.260} & \textbf{-0.254} & \textbf{-0.311} \\
H20 & \textbf{0.320} & \textbf{0.311} & \textbf{-0.329} & \textbf{-0.333} & \textemdash & \textemdash & \textemdash & \textemdash & \textbf{0.260} & \textbf{0.252} & \textbf{-0.242} & \textbf{-0.300} \\
H22 & \textbf{0.322} & \textbf{0.315} & \textbf{-0.328} & \textbf{-0.331} & -0.033 & -0.033 & 0.033 & 0.034 & \textbf{0.260} & \textbf{0.253} & \textbf{-0.245} & \textbf{-0.299} \\
H24 & \textbf{0.326} & \textbf{0.319} & \textbf{-0.331} & \textbf{-0.334} & -0.029 & -0.029 & 0.029 & 0.030 & \textbf{0.261} & \textbf{0.254} & \textbf{-0.249} & \textbf{-0.299} \\
H26 & \textbf{0.325} & \textbf{0.323} & \textbf{-0.332} & \textbf{-0.333} & -0.501 & -0.501 & 0.501 & 0.500 & \textbf{0.274} & \textbf{0.267} & \textbf{-0.263} & \textbf{-0.309} \\
H28 & \textbf{0.322} & \textbf{0.321} & \textbf{-0.331} & \textbf{-0.333} & -0.472 & -0.472 & 0.473 & 0.474 & \textbf{0.287} & \textbf{0.279} & \textbf{-0.281} & \textbf{-0.324} \\
H30 & \textbf{0.313} & \textbf{0.313} & \textbf{-0.321} & \textbf{-0.322} & -0.521 & -0.521 & 0.521 & 0.523 & \textbf{0.290} & \textbf{0.282} & \textbf{-0.288} & \textbf{-0.329} \\
H32 & \textbf{0.317} & \textbf{0.317} & \textbf{-0.324} & \textbf{-0.325} & -0.483 & -0.483 & 0.484 & 0.486 & \textbf{0.285} & \textbf{0.277} & \textbf{-0.284} & \textbf{-0.325} \\
\bottomrule
\end{tabular*}
\end{subtable}

\medskip

\begin{subtable}[t]{\linewidth}
\centering
\caption{\appOd{}}
\label{tab:rq2a_appod}

\begin{tabular*}{\linewidth}{@{\extracolsep{\fill}}l*{12}{r}@{}}
\toprule
& \multicolumn{4}{c}{Qwen} & \multicolumn{4}{c}{InternVL} & \multicolumn{4}{c}{LLaVA} \\
\cmidrule(lr){2-5}\cmidrule(lr){6-9}\cmidrule(lr){10-13}
History & MSP & PCS & DG & Ent. & MSP & PCS & DG & Ent. & MSP & PCS & DG & Ent. \\
\midrule
H02 & \textbf{0.190} & \textbf{0.188} & \textbf{-0.168} & \textbf{-0.165} & \textbf{0.015} & \textbf{0.014} & \textbf{-0.016} & \textbf{-0.017} & \textbf{0.086} & \textbf{0.081} & \textbf{-0.015} & \textbf{-0.070} \\
H04 & \textbf{0.207} & \textbf{0.200} & \textbf{-0.195} & \textbf{-0.208} & \textbf{0.019} & \textbf{0.018} & \textbf{-0.023} & \textbf{-0.027} & \textbf{0.238} & \textbf{0.233} & \textbf{-0.173} & \textbf{-0.216} \\
H06 & \textbf{0.252} & \textbf{0.243} & \textbf{-0.239} & \textbf{-0.254} & -0.096 & -0.099 & 0.109 & 0.115 & \textbf{0.234} & \textbf{0.231} & \textbf{-0.172} & \textbf{-0.214} \\
H08 & \textbf{0.247} & \textbf{0.237} & \textbf{-0.240} & \textbf{-0.260} & -0.178 & -0.179 & 0.205 & 0.219 & \textbf{0.251} & \textbf{0.246} & \textbf{-0.207} & \textbf{-0.260} \\
H10 & \textbf{0.272} & \textbf{0.261} & \textbf{-0.269} & \textbf{-0.289} & -0.126 & -0.126 & 0.134 & 0.135 & \textbf{0.231} & \textbf{0.224} & \textbf{-0.187} & \textbf{-0.251} \\
H12 & \textbf{0.297} & \textbf{0.285} & \textbf{-0.293} & \textbf{-0.312} & -0.128 & -0.128 & 0.142 & 0.145 & \textbf{0.220} & \textbf{0.215} & \textbf{-0.174} & \textbf{-0.232} \\
H14 & \textbf{0.314} & \textbf{0.303} & \textbf{-0.307} & \textbf{-0.323} & -0.035 & -0.037 & 0.032 & 0.029 & \textbf{0.225} & \textbf{0.219} & \textbf{-0.184} & \textbf{-0.244} \\
H16 & \textbf{0.331} & \textbf{0.318} & \textbf{-0.330} & \textbf{-0.348} & -0.029 & -0.030 & 0.030 & 0.031 & \textbf{0.234} & \textbf{0.229} & \textbf{-0.206} & \textbf{-0.257} \\
H18 & \textbf{0.324} & \textbf{0.313} & \textbf{-0.329} & \textbf{-0.345} & -0.135 & -0.135 & 0.157 & 0.163 & \textbf{0.215} & \textbf{0.209} & \textbf{-0.181} & \textbf{-0.244} \\
H20 & \textbf{0.322} & \textbf{0.311} & \textbf{-0.321} & \textbf{-0.335} & -0.129 & -0.129 & 0.145 & 0.149 & \textbf{0.209} & \textbf{0.204} & \textbf{-0.179} & \textbf{-0.234} \\
H22 & \textbf{0.313} & \textbf{0.304} & \textbf{-0.308} & \textbf{-0.321} & -0.114 & -0.114 & 0.126 & 0.136 & \textbf{0.211} & \textbf{0.205} & \textbf{-0.177} & \textbf{-0.235} \\
H24 & \textbf{0.329} & \textbf{0.319} & \textbf{-0.326} & \textbf{-0.341} & -0.126 & -0.126 & 0.142 & 0.146 & \textbf{0.229} & \textbf{0.224} & \textbf{-0.199} & \textbf{-0.249} \\
H26 & \textbf{0.316} & \textbf{0.307} & \textbf{-0.306} & \textbf{-0.318} & \textbf{0.112} & \textbf{0.111} & \textbf{-0.076} & \textbf{-0.036} & \textbf{0.227} & \textbf{0.222} & \textbf{-0.203} & \textbf{-0.252} \\
H28 & \textbf{0.324} & \textbf{0.314} & \textbf{-0.312} & \textbf{-0.324} & \textbf{0.128} & \textbf{0.127} & \textbf{-0.098} & \textbf{-0.063} & \textbf{0.234} & \textbf{0.229} & \textbf{-0.216} & \textbf{-0.259} \\
H30 & \textbf{0.341} & \textbf{0.333} & \textbf{-0.334} & \textbf{-0.341} & \textbf{0.107} & \textbf{0.106} & \textbf{-0.072} & \textbf{-0.032} & \textbf{0.242} & \textbf{0.237} & \textbf{-0.227} & \textbf{-0.269} \\
H32 & \textbf{0.332} & \textbf{0.325} & \textbf{-0.323} & \textbf{-0.329} & \textbf{0.091} & \textbf{0.090} & \textbf{-0.052} & \textbf{-0.014} & \textbf{0.228} & \textbf{0.223} & \textbf{-0.211} & \textbf{-0.258} \\
\bottomrule
\end{tabular*}
\end{subtable}

\par\smallskip
\begin{minipage}{\linewidth}
\footnotesize
\textit{Note.} Values are point-biserial correlations ($r_{\mathrm{pb}}$) computed over frame-level predictions pooled across the 25 evaluation bags, with correctness coded as 1 for a correct prediction and 0 for an incorrect prediction. MSP and PCS are confidence measures, for which the expected reliability direction is $r_{\mathrm{pb}}>0$; DG (DeepGini) and Ent. (entropy) are uncertainty measures, for which the expected direction is $r_{\mathrm{pb}}<0$. Bold indicates that the observed coefficient follows this expected direction; it does not denote statistical significance. \textemdash{} denotes an unavailable coefficient.
\end{minipage}

\end{table}

Thus, Qwen shows the most consistent relationship between prediction correctness and its confidence/uncertainty estimates: correct predictions are associated with higher confidence and lower uncertainty across all histories, whereas InternVL shows a less stable pattern that changes with input type and history. These correlations indicate whether confidence and uncertainty vary as expected between correct and incorrect predictions; they do not measure how closely the reported confidence values match the model's actual accuracy. Thus, Qwen is the most directionally consistent (correct predictions have higher confidence and lower uncertainty), while InternVL is sensitive to input type and history. These correlations assess association with correctness, not calibration.

\subsubsection{RQ2(b): Model Comparison}\label{sec:rq2b_model_comparison}

RQ2(b) compares the three VLMs within each input type and history using bag-level association and calibration metrics. Friedman tests assess overall model differences, followed by paired Wilcoxon signed-rank tests with Holm correction. Table~\ref{tab:rq2b_model_comparison} reports a model only when it significantly outperforms both competitors.
\begin{table}[tb]
\caption{RQ2(b) -- Models that significantly outperform both competitors for confidence, uncertainty, and calibration metrics within each input type and history.}
\label{tab:rq2b_model_comparison}
\centering
\scriptsize
\setlength{\tabcolsep}{1.0pt}
\begin{tabular*}{\linewidth}{@{\extracolsep{\fill}}ll*{16}{c}@{}}
\toprule
& & \multicolumn{16}{c}{History} \\
\cmidrule(lr){3-18}
Approach & Metric & H02 & H04 & H06 & H08 & H10 & H12 & H14 & H16 & H18 & H20 & H22 & H24 & H26 & H28 & H30 & H32 \\
\midrule
\appFr{} & MSP $r$        & \textbf{Q} & \textbf{Q} & \textbf{Q} & \textbf{Q} & \textemdash & \textemdash & \textbf{Q} & \textbf{Q} & \textemdash & \textemdash & \textemdash & \textemdash & \textemdash & \textemdash & \textemdash & \textemdash \\
& PCS $r$        & \textbf{Q} & \textbf{Q} & \textbf{Q} & \textbf{Q} & \textemdash & \textbf{Q} & \textbf{Q} & \textbf{Q} & \textemdash & \textemdash & \textemdash & \textemdash & \textemdash & \textemdash & \textemdash & \textemdash \\
& DeepGini $r$  & \textbf{Q} & \textbf{Q} & \textbf{Q} & \textbf{Q} & \textemdash & \textbf{Q} & \textbf{Q} & \textbf{Q} & \textemdash & \textemdash & \textemdash & \textemdash & \textemdash & \textemdash & \textemdash & \textemdash \\
& Entropy $r$    & \textbf{Q} & \textbf{Q} & \textbf{Q} & \textemdash & \textemdash & \textemdash & \textemdash & \textemdash & \textemdash & \textemdash & \textemdash & \textemdash & \textemdash & \textemdash & \textemdash & \textemdash \\
& ECE            & \textemdash & \textemdash & \textbf{Q} & \textbf{Q} & \textbf{Q} & \textbf{Q} & \textemdash & \textbf{Q} & \textbf{Q} & \textbf{Q} & \textbf{Q} & \textbf{Q} & \textbf{Q} & \textbf{Q} & \textbf{Q} & \textbf{Q} \\
& MCE            & \textemdash & \textemdash & \textemdash & \textemdash & \textemdash & \textemdash & \textemdash & \textemdash & \textemdash & \textemdash & \textemdash & \textemdash & \textemdash & \textbf{Q} & \textemdash & \textemdash \\
\addlinespace
\appOd{} & MSP $r$        & \textbf{Q} & \textbf{Q} & \textbf{Q} & \textbf{Q} & \textbf{Q} & \textbf{Q} & \textbf{Q} & \textbf{Q} & \textbf{Q} & \textbf{Q} & \textbf{Q} & \textbf{Q} & \textemdash & \textemdash & \textemdash & \textemdash \\
& PCS $r$        & \textbf{Q} & \textemdash & \textbf{Q} & \textbf{Q} & \textbf{Q} & \textbf{Q} & \textbf{Q} & \textbf{Q} & \textbf{Q} & \textbf{Q} & \textbf{Q} & \textbf{Q} & \textemdash & \textemdash & \textemdash & \textemdash \\
& DeepGini $r$  & \textbf{Q} & \textbf{Q} & \textbf{Q} & \textbf{Q} & \textbf{Q} & \textbf{Q} & \textbf{Q} & \textbf{Q} & \textbf{Q} & \textbf{Q} & \textbf{Q} & \textbf{Q} & \textemdash & \textemdash & \textemdash & \textemdash \\
& Entropy $r$    & \textbf{Q} & \textbf{Q} & \textbf{Q} & \textbf{Q} & \textbf{Q} & \textbf{Q} & \textbf{Q} & \textbf{Q} & \textbf{Q} & \textbf{Q} & \textbf{Q} & \textemdash & \textemdash & \textemdash & \textemdash & \textemdash \\
& ECE            & \textemdash & \textemdash & \textemdash & \textemdash & \textemdash & \textemdash & \textemdash & \textemdash & \textemdash & \textemdash & \textemdash & \textemdash & \textemdash & \textemdash & \textemdash & \textemdash \\
& MCE            & \textbf{Q} & \textemdash & \textbf{Q} & \textbf{Q} & \textbf{Q} & \textemdash & \textbf{Q} & \textemdash & \textbf{Q} & \textbf{Q} & \textbf{Q} & \textbf{Q} & \textemdash & \textemdash & \textemdash & \textbf{Q} \\
\bottomrule
\end{tabular*}
\par\smallskip
\begin{minipage}{\linewidth}
\footnotesize
\textit{Note.} Q denotes Qwen. A model is listed only when the corresponding model-comparison test is significant after Holm correction and the model significantly outperforms both competitors in Holm-adjusted pairwise Wilcoxon tests ($p_{\mathrm{Holm}}<0.05$). \textemdash{} indicates that no model meets this criterion; individual model pairs may still differ significantly. For MSP and PCS correlations, a stronger positive association with correctness is preferable; for DeepGini and entropy correlations, a stronger negative association is preferable. Lower ECE and MCE indicate better calibration.
\end{minipage}
\end{table}

Qwen is the only unique significant winner in Table~\ref{tab:rq2b_model_comparison}. Under \appFr{}, it wins MSP at 6 histories, PCS and DG at 7 each, entropy at 3, ECE at 13, and MCE only at H28. Under \appOd{}, Qwen wins MSP and DG at 12 histories, PCS and entropy at 11, and MCE at 10, while no unique ECE winner is identified. The association-metric advantage is concentrated before H26. Overall, Qwen is the most consistently supported model, although the advantage depends on metric, input type, and history. Complete Friedman and pairwise results are provided in the supplementary material~\cite{farooq2026vlmrepo}.

\subsubsection{RQ2(c): Overall Model Comparison}
\label{sec:rq2c_overall_model_comparison}

RQ2(c) compares the three VLMs across all input types, histories, and bags. The Friedman tests show significant differences among the models for all six reliability metrics after Holm correction. The corresponding Kendall's $W$ values indicate moderate differences for MSP, PCS, DG, and entropy ($W=0.400$--$0.437$), a stronger difference for ECE ($W=0.510$), and a smaller difference for MCE ($W=0.163$).

The pairwise Wilcoxon comparisons are also significant for every model pair and metric (Table~\ref{tab:rq2c_overall_model_comparison}).
\begin{table}[!tb]
\caption{RQ2(c) -- Aggregate pairwise model comparisons for confidence, uncertainty, and calibration across all evaluated configurations.}
\label{tab:rq2c_overall_model_comparison}
\centering
\small
\setlength{\tabcolsep}{9pt}
\begin{tabular}{lllrlrr}
\toprule
Metric & Model 1 & Model 2 & $r_{\mathrm{rb}}$ & Effect size & Better model & $p_{\mathrm{Holm}}$ \\
\midrule
MSP
& Qwen     & InternVL &  0.903 & Large  & Qwen  & $\mathbf{9.07\times10^{-89}}$ \\
& Qwen     & LLaVA    &  0.621 & Large  & Qwen  & $\mathbf{3.49\times10^{-43}}$ \\
& InternVL & LLaVA    & -0.780 & Large  & LLaVA & $\mathbf{1.23\times10^{-66}}$ \\
\addlinespace

PCS
& Qwen     & InternVL &  0.898 & Large  & Qwen  & $\mathbf{1.21\times10^{-87}}$ \\
& Qwen     & LLaVA    &  0.620 & Large  & Qwen  & $\mathbf{5.74\times10^{-43}}$ \\
& InternVL & LLaVA    & -0.771 & Large  & LLaVA & $\mathbf{3.09\times10^{-65}}$ \\
\addlinespace

DeepGini
& Qwen     & InternVL &  0.913 & Large  & Qwen  & $\mathbf{1.33\times10^{-90}}$ \\
& Qwen     & LLaVA    &  0.647 & Large  & Qwen  & $\mathbf{9.91\times10^{-47}}$ \\
& InternVL & LLaVA    & -0.782 & Large  & LLaVA & $\mathbf{4.30\times10^{-67}}$ \\
\addlinespace

Entropy
& Qwen     & InternVL &  0.925 & Large  & Qwen  & $\mathbf{5.80\times10^{-93}}$ \\
& Qwen     & LLaVA    &  0.411 & Medium & Qwen  & $\mathbf{7.51\times10^{-20}}$ \\
& InternVL & LLaVA    & -0.857 & Large  & LLaVA & $\mathbf{3.49\times10^{-80}}$ \\
\addlinespace

ECE
& Qwen     & InternVL &  0.996 & Large  & Qwen  & $\mathbf{3.88\times10^{-131}}$ \\
& Qwen     & LLaVA    &  0.344 & Medium & Qwen  & $\mathbf{3.40\times10^{-17}}$ \\
& InternVL & LLaVA    & -0.843 & Large  & LLaVA & $\mathbf{1.61\times10^{-94}}$ \\
\addlinespace

MCE
& Qwen     & InternVL &  0.680 & Large & Qwen  & $\mathbf{6.65\times10^{-62}}$ \\
& Qwen     & LLaVA    &  0.549 & Large & Qwen  & $\mathbf{6.04\times10^{-41}}$ \\
& InternVL & LLaVA    & -0.131 & Small & LLaVA & $\mathbf{1.28\times10^{-3}}$ \\
\midrule
\multicolumn{7}{c}{\textbf{Overall ordering for all metrics: Qwen $>$ LLaVA $>$ InternVL}} \\
\bottomrule
\end{tabular}

\par\smallskip
\begin{minipage}{\linewidth}
\footnotesize
\textit{Note.} $r_{\mathrm{rb}}$ denotes the matched-pairs rank-biserial correlation for Model~1 minus Model~2 after orienting each metric so that larger values indicate better reliability or calibration. Positive values therefore favor Model~1, whereas negative values favor Model~2. Effect-size magnitudes are interpreted using $|r_{\mathrm{rb}}|$ as small ($\geq0.11$), medium ($\geq0.28$), and large ($\geq0.43$). Bold $p_{\mathrm{Holm}}$ values indicate statistically significant pairwise differences after Holm correction. The pooled comparison summarizes repeated model evaluations across matched bags, approaches, and history lengths.
\end{minipage}
\end{table}
After orienting the metrics so that larger values indicate better reliability or calibration, all six metrics show the same overall ordering: Qwen $>$ LLaVA $>$ InternVL. The strength of the pairwise differences varies across metrics. For example, the LLaVA--InternVL difference is small for MCE ($|r_{\mathrm{rb}}|=0.131$), but considerably larger for entropy ($0.857$) and ECE ($0.843$). Thus, the overall ordering is consistent, although the magnitude of the separation between models depends on the reliability measure.

\answerRQ{
\textbf{Answer to RQ2.}
Qwen shows the most reliable confidence and uncertainty behavior across the evaluated settings. Its confidence measures consistently increase with correct predictions, while its uncertainty measures decrease, whereas InternVL shows more history- and input-type-dependent behavior. The within-setting comparisons also identify Qwen as the most consistently supported model across confidence, uncertainty, and calibration measures. In the aggregate comparison, all six metrics give the same ordering: $\mathrm{Qwen}$ $>$ $\mathrm{LLaVA}$ $>$ $\mathrm{InternVL}$. Qwen provides the strongest evidence that its reported confidence and uncertainty reflect prediction reliability.
}

\subsection{RQ3: Effect of History Length}\label{sec:rq3}

In this RQ, we examined how the length of the visual history affects classification performance for each model--input type configuration using bag-level Accuracy and Macro-F1 scores across the 16 histories (H02--H32). For each configuration, a Friedman test first assesses whether performance differs across histories. The six omnibus tests for each metric are Holm-corrected. When the corrected Friedman test is significant, the 120 possible history pairs are compared using paired Wilcoxon signed-rank tests with Holm correction. Table~\ref{tab:rq3_history_effect} summarizes the omnibus tests and the histories accumulating the largest numbers of significant pairwise wins.

\begin{table}[!tb]
\caption{RQ3 -- Effect of history length on classification performance.}
\label{tab:rq3_history_effect}
\centering
\scriptsize
\setlength{\tabcolsep}{2.5pt}

\begin{subtable}[t]{\linewidth}
\centering
\caption{Accuracy}
\label{tab:rq3_accuracy}

\begin{tabular}{@{}llccrrrclll@{}}
\toprule
Model & Input Type & Bags & Hist. & $\chi_F^2$ & $p_{\mathrm{Holm}}$ & $W$ &
Sig. pairs & Most wins & Second tier & Third tier \\
\midrule
Qwen     & \appFr{} & 25 & 16 & 46.832  & $<0.001$ & 0.125 & 0/120  & \textemdash & \textemdash & \textemdash \\
Qwen     & \appOd{} & 25 & 16 & 105.041 & $<0.001$ & 0.280 & 14/120 & H32 (4) & H30 (3) & H26, H28 (2 each) \\
InternVL & \appFr{} & 25 & 16 & 117.504 & $<0.001$ & 0.313 & 53/120 & H02, H14, H16 (12 each) & H04 (11) & H06 (3) \\
InternVL & \appOd{} & 25 & 16 & 78.461  & $<0.001$ & 0.209 & 26/120 & H14 (5) & H02, H22 (4 each) & H30 (3) \\
LLaVA    & \appFr{} & 25 & 16 & 67.202  & $<0.001$ & 0.179 & 4/120  & H18, H20 (2 each) & \textemdash & \textemdash \\
LLaVA    & \appOd{} & 25 & 16 & 49.213  & $<0.001$ & 0.131 & 1/120  & H16 (1) & \textemdash & \textemdash \\
\bottomrule
\end{tabular}
\end{subtable}

\medskip

\begin{subtable}[t]{\linewidth}
\centering
\caption{Macro-F1}
\label{tab:rq3_macrof1}

\begin{tabular}{@{}llccrrrclll@{}}
\toprule
Model & Input Type & Bags & Hist. & $\chi_F^2$ & $p_{\mathrm{Holm}}$ & $W$ &
Sig. pairs & Most wins & Second tier & Third tier \\
\midrule
Qwen     & \appFr{} & 25 & 16 & 22.123  & 0.418    & 0.059 & Not performed & \textemdash & \textemdash & \textemdash \\
Qwen     & \appOd{} & 25 & 16 & 19.353  & 0.595    & 0.052 & Not performed & \textemdash & \textemdash & \textemdash \\
InternVL & \appFr{} & 25 & 16 & 143.629 & $<0.001$ & 0.383 & 33/120 & H02 (12) & H14 (7) & H06 (5) \\
InternVL & \appOd{} & 25 & 16 & 82.977  & $<0.001$ & 0.221 & 20/120 &
\shortstack[l]{H02, H04, H14, H16, H18,\\H20, H22, H24, H30 (2 each)} &
H10, H12 (1 each) & \textemdash \\
LLaVA    & \appFr{} & 25 & 16 & 17.044  & 0.625    & 0.045 & Not performed & \textemdash & \textemdash & \textemdash \\
LLaVA    & \appOd{} & 25 & 16 & 17.108  & 0.625    & 0.046 & Not performed & \textemdash & \textemdash & \textemdash \\
\bottomrule
\end{tabular}
\end{subtable}

\par\smallskip
\begin{minipage}{\linewidth}
\footnotesize
\textit{Note.} ``Sig. pairs'' reports the number of Holm-significant paired Wilcoxon comparisons out of the 120 possible history pairs. Pairwise comparisons are performed only when the corresponding Holm-corrected Friedman test is significant. The win tiers summarize histories according to the number of other histories they significantly outperform. $W$ denotes Kendall's $W$ effect size.
\end{minipage}

\end{table}

\begin{figure}[!tb]
\centering
\includegraphics[width=\linewidth]{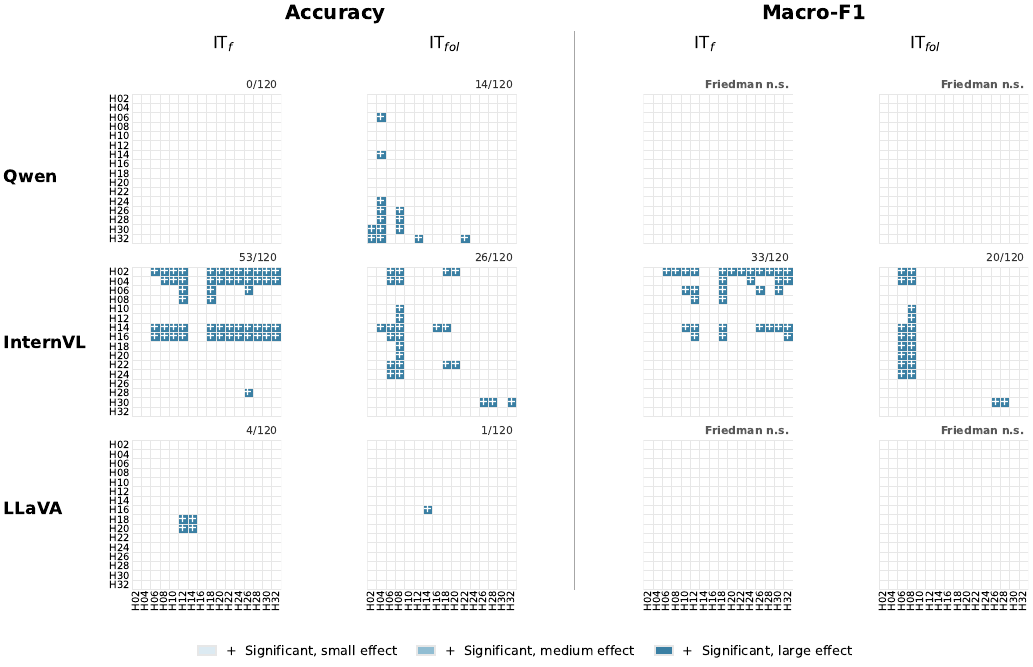}
\caption{RQ3 -- Pairwise history-length comparisons for Accuracy and Macro-F1. Within each matrix, a ``+'' indicates that the row history significantly outperforms the column history after Holm correction. Cell shading represents the magnitude of the matched-pairs rank-biserial effect size ($r_{\mathrm{rb}}$): small, medium, or large. Blank cells indicate no significant pairwise difference. The value in the upper-right corner of each matrix reports the number of significant pairs among the 120 possible history comparisons. \textit{Friedman n.s.} denotes a nonsignificant Holm-corrected Friedman test, for which post-hoc pairwise comparisons were not performed.}
\Description{}
\label{fig:rq3_pairwise_history}
\end{figure}

\begin{figure}[!tb]
\centering
\begin{subfigure}[t]{0.49\linewidth}

\centering
\includegraphics[width=\linewidth]{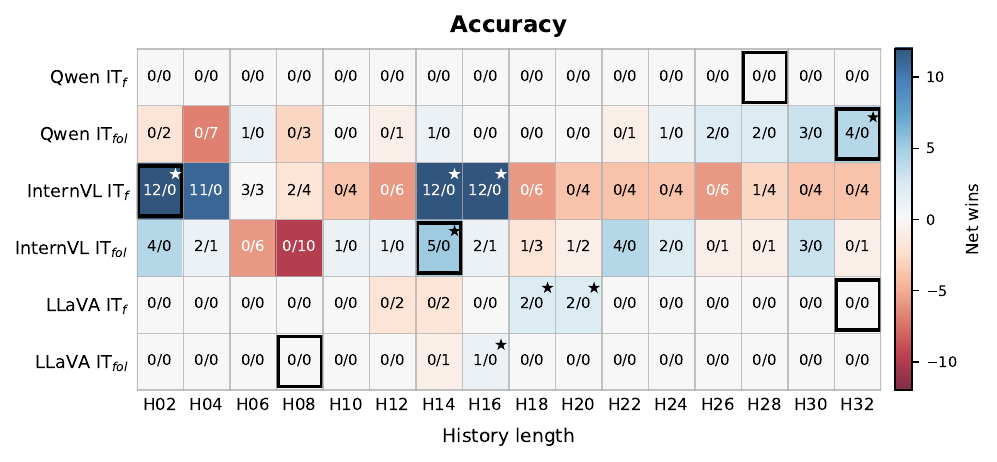}
\caption{Accuracy}
\label{fig:rq3_history_evidence_accuracy}
\end{subfigure}
\hfill
\begin{subfigure}[t]{0.49\linewidth}
\centering
\includegraphics[width=\linewidth]{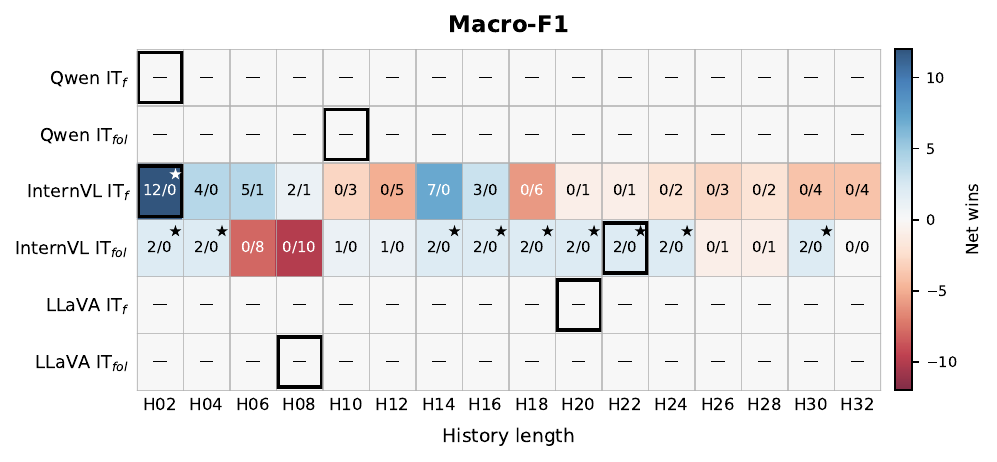}
\caption{Macro-F1}
\label{fig:rq3_history_evidence_macrof1}
\end{subfigure}
\caption{RQ3 -- History-selection evidence summarized from the Holm-adjusted pairwise comparisons. Each cell reports significant wins/losses for a history; the fill represents net wins, a star marks histories with the largest number of significant wins within a model--input type configuration, and the outlined cell marks the highest mean performance. A dash indicates that the omnibus Friedman test is nonsignificant and post-hoc comparisons are therefore not performed.}
\Description{}
\label{fig:rq3_history_evidence}
\end{figure}

Figure~\ref{fig:rq3_pairwise_history} provides the complete pairwise pattern, while Figure~\ref{fig:rq3_history_evidence} summarizes the corresponding win--loss evidence.

\subsubsection{Accuracy}\label{sec:rq3_accuracy}

History length significantly affects Accuracy for all six model--input type configurations after Holm correction ($p_{\mathrm{Holm}}<0.001$), although the strength and localization of the effect vary considerably (Table~\ref{tab:rq3_accuracy}). Kendall's $W$ ranges from 0.125 for Qwen--\appFr{} to 0.313 for InternVL--\appFr{}. Despite the significant omnibus result, Qwen--\appFr{} has no pairwise comparison that survives Holm correction, indicating that its Accuracy is comparatively stable across individual history lengths. In contrast, Qwen--\appOd{} shows 14 significant pairs, with the strongest evidence favoring longer histories: H32 records four significant wins and H30 records three.

InternVL is substantially more history-sensitive. Under \appFr{}, 53 of 120 pairwise comparisons are significant, with H02, H14, and H16 each outperforming 12 other histories; under \appOd{}, 26 pairs are significant and H14 has the largest number of wins (five). LLaVA shows much weaker pairwise separation despite significant omnibus tests: only four pairs are significant under \appFr{} and one under \appOd{}. Thus, Accuracy does not support a single history length across all configurations; the preferred history is strongly model- and input type-dependent.

\subsubsection{Macro-F1}\label{sec:rq3_macrof1}

The history effect is less general for Macro-F1 (Table~\ref{tab:rq3_macrof1}). Neither Qwen nor LLaVA shows a significant Friedman result under either input type after Holm correction ($p_{\mathrm{Holm}}=0.418$--$0.625$); consequently, pairwise post-hoc tests are not performed for these four configurations. This indicates no statistical evidence that changing history length systematically alters their Macro-F1 over the evaluated range.

InternVL is the exception. Under \appFr{}, the history effect is the strongest observed in RQ3 ($\chi_F^2=143.629$, $W=0.383$), with 33 significant pairwise differences. H02 has the clearest advantage, significantly outperforming 12 other histories, followed by H14 with seven wins and H06 with five. Under \appOd{}, the effect remains significant but weaker ($W=0.221$), with 20 significant pairs. Here the significant wins are distributed across several histories rather than concentrated on a single history: H2, H4, H14, H16, H18, H20, H22, H24, H30 each significantly outperforms two other histories. 
while H10 and H12 each outperform once. 

Overall, RQ3 shows that longer visual histories are not uniformly better. For Accuracy, Qwen--\appOd{} provides the clearest evidence in favor of long histories, particularly H30--H32, whereas Qwen--\appFr{} remains pairwise stable despite a significant omnibus effect. InternVL is highly history-sensitive, but its preferred histories depend on the input type: H02/H14/H16 dominate under \appFr{}, while H14 is strongest under \appOd{}. LLaVA exhibits only sparse pairwise differences. For Macro-F1, history length is statistically relevant only for InternVL, with H02 providing the strongest evidence under \appFr{} and no unique optimum emerging under \appOd{}. Accordingly, the results do not justify a single globally optimal history length; history selection should be conditioned on both the VLM and the input type, and should consider Accuracy and class-balanced Macro-F1 jointly.

\answerRQ{
\textbf{Answer to RQ3.}
History length affects classification performance differently across VLMs and input types. Longer histories are most clearly supported for Qwen with \appOd{}, particularly H30--H32, whereas InternVL often favors shorter or intermediate histories, especially H02 under \appFr{}. LLaVA shows only limited pairwise differences. For Macro-F1, significant history differences are observed only for InternVL. Overall, no single history length is optimal across all configurations; history selection should therefore consider the VLM, input type, and evaluation metric jointly.
}

\subsection{RQ4: Early-Warning Capability}\label{sec:rq4}

In this RQ, we are interested in whether the evaluated VLM configurations can provide early and actionable warnings before unsafe events.

RQ4 examines whether history settings supported by the preceding performance analyses also provide useful early warnings. To keep history selection independent of the warning outcomes, the configurations are selected before examining RQ4 results. For InternVL, we retain histories with at least one Holm-significant pairwise win in both Accuracy and Macro-F1 in RQ3. This yields 17 InternVL configurations. We additionally evaluate Qwen because it provides the strongest overall performance in RQ1 and RQ2; for Qwen, we retain the seven histories with at least one Holm-significant Accuracy win in RQ3. Macro-F1 post-hoc comparisons are not available for Qwen because its corresponding Friedman tests are nonsignificant. The resulting RQ4 analysis therefore contains 24 model--input type--history configurations.

Warning timing is evaluated using the lead-time and stopping-time definitions in Eqs.~\eqref{eq:lead-time}--\eqref{eq:warning-outcome}. For each configuration, the 86 unsafe events are classified as actionable, too early, too late, or missed. The allowed warning interval is varied over $A\in\{1,2,3,4,5\}$\,s.

Table~\ref{tab:rq4_selected_warning_outcomes} shows that warning coverage and actionable-warning performance do not identify the same configuration as best.
\begin{table}[!tb]
\caption{RQ4 -- Early-warning outcomes for histories selected from the preceding performance analyses. Each $A$ column reports actionable/too-early warnings (AW/TE) out of 86 unsafe events. Acc. and MF1 denote the numbers of significant history wins in RQ3; N.P. indicates that Macro-F1 post-hoc comparisons were not performed. TL, M, and Cov. denote too late, missed, and warning coverage, respectively.}
\label{tab:rq4_selected_warning_outcomes}
\centering
\scriptsize
\setlength{\tabcolsep}{2.0pt}
\resizebox{\linewidth}{!}{%
\begin{tabular}{@{}lllrrcccccclll@{}}
\toprule
Model & Input Type & History & Acc. & MF1 & \multicolumn{5}{c}{$A$} & TL & M & Cov. \\
\cmidrule{6-10}
& & & & & $1$ & $2$ & $3$ & $4$ & $5$\\
\midrule
Qwen & \appOd{} & H06 & 1 & N.P.
& 7/59 & 10/56 & 12/54 & 14/52 & 15/51
& 12 (13.95\%) & 8 (9.30\%) & 78 (90.70\%) \\

Qwen & \appOd{} & H14 & 1 & N.P.
& 4/57 & 5/56 & 8/53 & 11/50 & 12/49
& 10 (11.63\%) & 15 (17.44\%) & 71 (82.56\%) \\

Qwen & \appOd{} & H24 & 1 & N.P.
& 4/52 & 7/49 & 10/46 & 13/43 & 13/43
& 8 (9.30\%) & 22 (25.58\%) & 64 (74.42\%) \\

Qwen & \appOd{} & H26 & 2 & N.P.
& 5/54 & 8/51 & 11/48 & 13/46 & 13/46
& 10 (11.63\%) & 17 (19.77\%) & 69 (80.23\%) \\

Qwen & \appOd{} & H28 & 2 & N.P.
& 6/53 & 8/51 & 11/48 & 15/44 & 15/44
& 9 (10.47\%) & 18 (20.93\%) & 68 (79.07\%) \\

Qwen & \appOd{} & H30 & 3 & N.P.
& 4/54 & 6/52 & 12/46 & 14/44 & 14/44
& 11 (12.79\%) & 17 (19.77\%) & 69 (80.23\%) \\

Qwen & \appOd{} & H32 & 4 & N.P.
& 6/50 & 7/49 & 11/45 & 14/42 & 14/42
& 10 (11.63\%) & 20 (23.26\%) & 66 (76.74\%) \\

\addlinespace

InternVL & \appFr{} & H02 & 12 & 12
& 9/59 & 11/57 & 14/54 & 16/52 & 19/49
& 11 (12.79\%) & 7 (8.14\%) & 79 (91.86\%) \\

InternVL & \appFr{} & H04 & 11 & 4
& 4/58 & 8/54 & 10/52 & 11/51 & 15/47
& 12 (13.95\%) & 12 (13.95\%) & 74 (86.05\%) \\

InternVL & \appFr{} & H06 & 3 & 5
& 6/63 & 10/59 & 14/55 & 16/53 & 18/51
& 15 (17.44\%) & 2 (2.33\%) & 84 (97.67\%) \\

InternVL & \appFr{} & H08 & 2 & 2
& 6/63 & 10/59 & 14/55 & 16/53 & 18/51
& 14 (16.28\%) & 3 (3.49\%) & 83 (96.51\%) \\

InternVL & \appFr{} & H14 & 12 & 7
& 5/58 & 8/55 & 10/53 & 12/51 & 14/49
& 14 (16.28\%) & 9 (10.47\%) & 77 (89.53\%) \\

InternVL & \appFr{} & H16 & 12 & 3
& 5/58 & 8/55 & 10/53 & 12/51 & 14/49
& 13 (15.12\%) & 10 (11.63\%) & 76 (88.37\%) \\

\addlinespace

InternVL & \appOd{} & H02 & 4 & 2
& 6/62 & 9/59 & 14/54 & 15/53 & 17/51
& 11 (12.79\%) & 7 (8.14\%) & 79 (91.86\%) \\

InternVL & \appOd{} & H04 & 2 & 2
& 6/62 & 8/60 & 13/55 & 15/53 & 18/50
& 12 (13.95\%) & 6 (6.98\%) & 80 (93.02\%) \\

InternVL & \appOd{} & H10 & 1 & 1
& 6/63 & 10/59 & 14/55 & 16/53 & 18/51
& 12 (13.95\%) & 5 (5.81\%) & 81 (94.19\%) \\

InternVL & \appOd{} & H12 & 1 & 1
& 5/63 & 10/58 & 13/55 & 15/53 & 17/51
& 15 (17.44\%) & 3 (3.49\%) & 83 (96.51\%) \\

InternVL & \appOd{} & H14 & 5 & 2
& 7/59 & 9/57 & 13/53 & 14/52 & 16/50
& 12 (13.95\%) & 8 (9.30\%) & 78 (90.70\%) \\

InternVL & \appOd{} & H16 & 2 & 2
& 5/61 & 8/58 & 12/54 & 13/53 & 16/50
& 11 (12.79\%) & 9 (10.47\%) & 77 (89.53\%) \\

InternVL & \appOd{} & H18 & 1 & 2
& 6/62 & 9/59 & 14/54 & 15/53 & 17/51
& 15 (17.44\%) & 3 (3.49\%) & 83 (96.51\%) \\

InternVL & \appOd{} & H20 & 1 & 2
& 6/62 & 10/58 & 13/55 & 15/53 & 17/51
& 15 (17.44\%) & 3 (3.49\%) & 83 (96.51\%) \\

InternVL & \appOd{} & H22 & 4 & 2
& 7/61 & 10/58 & 14/54 & 15/53 & 17/51
& 15 (17.44\%) & 3 (3.49\%) & 83 (96.51\%) \\

InternVL & \appOd{} & H24 & 2 & 2
& 7/61 & 10/58 & 13/55 & 15/53 & 17/51
& 15 (17.44\%) & 3 (3.49\%) & 83 (96.51\%) \\

InternVL & \appOd{} & H30 & 3 & 2
& 5/63 & 8/60 & 13/55 & 15/53 & 17/51
& 16 (18.60\%) & 2 (2.33\%) & 84 (97.67\%) \\

\bottomrule
\end{tabular}%
}
\end{table}
Across the 24 selected configurations, InternVL--\appFr{} H06 and InternVL--\appOd{} H30 achieve the highest warning coverage, 84/86 (97.67\%), and each misses only two unsafe events. In contrast, InternVL--\appFr{} H02 provides the largest number of actionable warnings at every tested value of $A$, increasing from 9/86 (10.47\%) at $A=1$\,s to 19/86 (22.09\%) at $A=5$\,s. At $A=5$\,s, InternVL--\appFr{} H06 provides slightly fewer actionable warnings, 18/86 (20.93\%), despite its higher overall coverage.

The selected Qwen configurations show a different pattern. Qwen--\appOd{} H06 provides the highest warning coverage among the selected Qwen histories, 78/86 (90.70\%), while also producing 7/86 (8.14\%) actionable warnings at $A=1$\,s and 15/86 (17.44\%) at $A=5$\,s. Qwen--\appOd{} H28 also reaches 15/86 (17.44\%) actionable warnings at $A=5$\,s, but its warning coverage is lower at 68/86 (79.07\%). The longest selected Qwen history, H32, records the largest number of significant Accuracy wins in RQ3 (four), but its warning coverage is only 66/86 (76.74\%). Thus, stronger classification evidence for a history does not necessarily translate into stronger early-warning coverage or timing.

Too-early warnings remain the most common outcome across both models. As $A$ increases, the accepted warning interval becomes wider, so some warnings previously classified as too early become actionable. For InternVL--\appFr{} H02, too-early warnings decrease from 59/86 (68.60\%) at $A=1$\,s to 49/86 (56.98\%) at $A=5$\,s, while actionable warnings increase from 9 to 19. A similar pattern is observed for Qwen--\appOd{} H06, where too-early warnings decrease from 59/86 (68.60\%) to 51/86 (59.30\%), while actionable warnings increase from 7 to 15. In contrast, too-late and missed counts remain unchanged for a fixed configuration because their definitions do not depend on $A$.

Overall, the histories favored by the preceding classification analyses do not produce a single best early-warning setting. InternVL--\appFr{} H02 provides the strongest actionable-warning performance, whereas InternVL--\appFr{} H06 and InternVL--\appOd{} H30 provide the highest warning coverage. Among the selected Qwen histories, \appOd{} H06 provides the strongest coverage and competitive actionable-warning performance. These results show that history selection should consider both classification performance and warning timing: a history that performs well for classification does not necessarily generate warnings at the most useful time.

\answerRQ{
\textbf{Answer to RQ4.}
The configurations with the strongest classification evidence do not necessarily provide the best early-warning behavior. InternVL--\appFr{} H06 and InternVL--\appOd{} H30 achieve the highest warning coverage (97.67\%), whereas InternVL--\appFr{} H02 provides the highest actionable-warning rate (22.09\% at $A=5$\,s). Among the selected Qwen configurations, \appOd{} H06 provides the highest warning coverage (90.70\%) and competitive actionable-warning performance. Overall, early-warning quality depends not only on whether a risk is detected, but also on when the warning is produced.
}
\section{Discussion}
\label{sec:Discussion}
This section discusses the main implications of our findings for using VLMs in robot-safety monitoring and highlights how the results can inform the design of future VLM-based robot digital twins.
\subsection{From Unsafe-State Detection to Early Warning}
\label{sec:discussion-anticipatory-warning}

A key implication of our results is that recognizing an already
\emph{unsafe} state should not be the only objective of VLM-based robot-safety monitoring. For preventive action, identifying a \emph{potentially unsafe} situation before the unsafe event occurs is more useful because it provides the robot with time to react. This distinction is particularly important in our results: direct recall of the \emph{unsafe} class remains very low across the evaluated configurations, whereas several VLM configurations generate warnings before a large proportion of unsafe events. Thus, weak recognition at the exact unsafe observation does not necessarily mean that the VLM provides no useful safety information. A model may detect visual or motion patterns indicating that risk is developing before the final unsafe state is reached. One possible explanation is that observations recorded at the moment of collision or near-contact may contain rapid camera motion, occlusion, or image blur, which could make the final unsafe state more difficult to classify visually. Nevertheless, the early-warning results show that useful evidence can already be present in preceding observations. This supports using the \emph{potentially unsafe} state as an anticipatory warning signal rather than relying only on recognition after the situation has already become unsafe.

\subsection{Prediction Reliability for Safety Decisions}
\label{sec:discussion-reliability}

Prediction reliability adds another dimension to this interpretation. Qwen provides the clearest evidence that confidence and uncertainty can help distinguish reliable from less reliable predictions: correct predictions consistently have higher confidence and lower uncertainty across the evaluated histories. InternVL, in contrast, provides stronger early-warning behavior but less stable confidence and uncertainty relationships across input types and histories. This suggests that early-warning capability and prediction reliability are complementary properties rather than the same property.

Therefore, confidence and uncertainty could be used together with the predicted safety state and warning lead time when deciding whether the system should continue monitoring, slow the robot, stop it, or request another preventive action.

\subsection{Temporal Context and Computational Trade-offs}
\label{sec:discussion-temporal-context}

The evaluation of multiple history lengths also has an important practical implication. A fixed long contextual memory is not always necessary and can substantially increase GPU memory, computation, and inference latency. Our preliminary experiments showed that increasing the number of input images increases memory requirements and can lead to out-of-memory failures, with H32 representing the largest history that could be evaluated consistently under our available resources.

The results further show that longer histories are not uniformly better. For example, InternVL with \appFr{02} provides the strongest actionable-warning result and \appFr{06} provides the highest visual-only warning coverage, showing that useful early-warning behavior can be obtained with relatively short histories. In contrast, longer histories such as H30--H32 receive stronger support for Qwen with \appOd{} when current-state classification is the objective.

These findings make the evaluated configurations practically useful rather than merely experimental combinations. Instead of maintaining the largest possible contextual history for every prediction, a robot digital twin can select a history according to the required function and available computational resources. A short-history configuration can reduce the number of images processed, GPU-memory usage, and computational cost when it already provides the desired early-warning behavior, while a longer history can be retained when the additional temporal context provides measurable classification benefits.
The configuration therefore represents a trade-off between the safety information required by the robot and the computational resources needed to produce that information.

\subsection{Practical Configuration Implications}
\label{sec:discussion-configurations}

Taken together, the results suggest that a future VLM-based robot digital twin should not be designed around one universally optimal model and history length. If timely visual-only warning is the priority, the present results support investigating InternVL--\appFr{02} for actionable warning or InternVL--\appFr{06} when warning coverage is prioritized. If odometry and LiDAR are available and high warning coverage is required, InternVL--\appOd{30} is the strongest evaluated configuration.

For current-state assessment with comparatively reliable confidence and uncertainty information, Qwen is better supported by the present results, with longer \appOd{} histories receiving stronger classification support. These are empirical configuration choices from the present study rather than universal prescriptions, but they provide concrete starting points for implementing and evaluating a live robot digital twin.

\section{Threats to Validity}
\label{sec:threats}

Several threats may affect the interpretation and generalizability of the results.

\subsubsection*{\bf Construct validity}
The safety states \emph{safe}, \emph{potentially unsafe}, and \emph{unsafe} are labeled using the sensing information available in each case study. Consequently, the numerical thresholds used to construct the ground truth are not identical across HuRoN, PAL, and CrowdBot. For HuRoN, the labels combine bumper/contact and proximity information with the closest pedestrian distance, using 0.6\,m and 1.2\,m as the unsafe and potentially unsafe boundaries; the choice of safety margins is guided by prior work on minimum separation distance, and the fisheye videos are used for manual verification. For PAL, the labels are derived from the minimum valid front-LiDAR distance in each 2\,s window, with 0.35\,m and 0.80\,m thresholds selected with reference to the physical scale of the TIAGo OMNI Base. CrowdBot follows the distance thresholds used in its preprocessing procedure, namely 0.5\,m and 1.2\,m. These case-specific definitions preserve the sensing characteristics and physical scale of each robot, but alternative thresholds could change the class distribution and, therefore, the measured performance. They also primarily represent distance- and contact-related hazards and may not cover every hazardous situation encountered in mobile-robot navigation. To avoid directly encoding these ground-truth rules into the VLMs, the numerical thresholds are not included in the prompts; instead, all models receive the same general descriptions of the three safety states. Nevertheless, because the models are evaluated zero-shot, alternative prompt formulations could produce different predictions.

The selected evaluation measures introduce additional construct-validity considerations. Accuracy can be influenced by the dominant \emph{safe} class, whereas Macro-F1 gives equal importance to all three classes; we therefore interpret these metrics together with per-class precision, recall, and F1. Similarly, MSP, PCS, entropy, and DG quantify properties of the predicted class-probability distribution and should not be interpreted as direct measures of physical danger, while ECE and MCE assess calibration and can depend on the chosen confidence-binning procedure. The early warning outcomes are also based on an operational definition. The minimum required warning time depends on the recorded robot speed together with the assumed reaction time, braking deceleration, and safety margin. In addition, the allowed warning interval $A$ is an evaluation parameter that determines how far before the minimum required warning time a prediction can still be considered actionable. Rather than fixing this parameter to a single value, we evaluate $A\in\{1,$ $2,$ $3,$ $4,$ $5\}$\,s and report \emph{actionable}, \emph{too early}, \emph{too late}, and \emph{missed} outcomes for each setting. This reduces dependence on one arbitrarily selected warning interval, although different stopping assumptions or values of $A$ could change the resulting early-warning
rates.

\subsubsection*{\bf Internal validity}
The experiments are performed on previously recorded robot trajectories. Therefore, every evaluated model receives observations from the same underlying robot runs, and its predictions cannot change the subsequent trajectory. This supports controlled comparison between configurations, but it does not capture possible interactions between a warning and the robot's subsequent behavior.

The same evaluation pipeline, safety-class formulation, and zero-shot setting are used for all three VLMs to reduce differences caused by the experimental procedure. The comparison between \appFr{} and \appOd{} also keeps the visual history unchanged, so that the main difference is the addition of odometry and LiDAR information. Nevertheless, differences between the internal architectures and probability outputs of the VLMs may affect how confidently they express their predictions.

History lengths are restricted to H02--H32. A preliminary study showed that larger histories substantially increased memory requirements and could lead to out-of-memory failures under the available computational resources. Therefore, the study cannot determine whether histories longer than H32 would improve or degrade performance. The selected range should be interpreted as the feasible range evaluated in this work rather than as a complete exploration of temporal context.

\subsubsection*{\bf Conclusion validity}
The safety classes are imbalanced, with considerably fewer \emph{unsafe} observations than \emph{safe} observations. This reduces the amount of evidence available for estimating unsafe class performance and is particularly important when interpreting the low unsafe recall observed for several configurations. For this reason, conclusions are not based on Accuracy alone and also consider Macro-F1 and per-class measures.

For the main statistical comparisons, performance is computed at the bag level rather than treating all individual frames as independent observations. This reduces the risk of artificially increasing the effective sample size because neighboring frames within the same recorded trajectory are temporally related. Paired tests are used where the same bags are evaluated under multiple conditions, and Holm correction is applied when multiple hypotheses are tested. For comparisons involving more than two matched conditions, Friedman tests are used before post-hoc Wilcoxon comparisons. Effect sizes are reported together with significance tests to distinguish statistical evidence from the magnitude of the observed difference.

The frame level analysis relating confidence and uncertainty to prediction correctness is an exception because it explicitly studies whether these values differ between correct and incorrect predictions. Since adjacent frames can be temporally dependent, these correlations should be interpreted as descriptive associations with correctness rather than as evidence that every frame provides an independent observation.

The early warning analysis contains 86 unsafe events for each selected configuration. Although this permits event-level comparison, the number of unsafe events is substantially smaller than the number of individual frames. Differences of only a few events can therefore change the reported warning percentages. The early-warning results should consequently be interpreted together with their underlying event counts.

\subsubsection*{\bf External validity}
The study evaluates three open-weight, general-purpose VLMs at a comparable 7--8B scale, two input types, and 16 history lengths across three mobile robot case studies. Using multiple datasets and robot settings provides broader evidence than evaluation on a single recorded environment, but the results cannot be assumed to apply to all robots, sensors, environments, VLM families, or safety tasks. In particular, larger or smaller VLMs, task-specific models, different camera configurations, different LiDAR systems, alternative prompts, or fine-tuned models may behave differently.

The practical configuration implications are therefore limited to the configurations evaluated in this study rather than representing universal rules. Future work should evaluate the selected configurations using live sensor streams and investigate how VLM-generated warnings affect actual robot actions, such as slowing down, stopping, or changing the navigation path.
\section{Conclusion and Future Work}
\label{sec:conclusion}

This study evaluated the potential of general-purpose VLMs for robot safety monitoring across three robotic case studies. We compared three VLMs, two input types, and 16 visual history settings, and analyzed classification performance, prediction reliability, the effect of temporal history, and early warning capability. The results show that VLMs can provide useful safety-related information and can generate warnings before unsafe events. Qwen provides the strongest overall classification and reliability results, particularly when visual history is enriched with odometry and LiDAR information. InternVL, in contrast, shows strong event-level early-warning capability, including effective visual-only configurations. These findings indicate that the most suitable VLM configuration depends on the intended safety objective rather than on classification performance alone.

As future work, we plan to develop a robot digital twin in which a VLM acts as the perception and reasoning module for safety monitoring. The digital twin will build on the configurations investigated in this study, including the choice of VLM, visual-history length, and input type, and will integrate camera observations with robot-state information such as odometry and LiDAR when appropriate. The goal is to move from the current offline evaluation toward an online system that continuously monitors the robot, estimates safety states and uncertainty, and provides early warnings before unsafe situations develop. We also plan to investigate closed-loop evaluation, improved handling of rare unsafe situations, and how VLM-generated warnings can be combined with conventional robot safety mechanisms for preventive action.

\begin{acks}
This work is supported by the InnoGuard Doctoral Network under the Marie Skłodowska-Curie Actions of the European Commission, Grant Agreement No. 101169233. Shaukat Ali, Paolo Arcaini, and Thomas Peyrucain are supported by the RoboSAPIENS project funded by the European Commission’s Horizon Europe programme under grant agreement number 101133807. Aitor Arrieta is part of the Software and Systems Engineering research group of Mondragon Unibertsitatea (IT1919-26), supported by the Department of Education, Universities and Research of the Basque Country. The authors acknowledge Simula Research Laboratory and the EX3 computing infrastructure for providing computational resources used in the experiments.
\end{acks}

\section*{Use of Generative AI}
\label{sec:genai_use}

OpenAI ChatGPT was used as a programming and workflow-support tool during the implementation and analysis of this study. The research questions, ground-truth definitions, experimental configurations, evaluation metrics, statistical procedures, and interpretation of the results were determined by the authors.

The experimental workflow was initially developed and tested by the authors on smaller data subsets. ChatGPT was then used to help scale the existing Python and Jupyter code to the full experimental setting, including adapting directory structures, configuration files, and scripts for multiple models, input types, history lengths, and case studies. It also assisted with migrating the VLM workflow to the EX3 GPU cluster, including SLURM batch scripts, environment configuration, and data synchronization.

For the data analysis and result preparation, the statistical procedures were selected and initially tested by the authors. ChatGPT was used to help adapt the analysis code for repeated execution over the complete results, including Friedman and paired Wilcoxon tests, multiple-comparison correction, CSV aggregation, and generation of LaTeX tables. All code, outputs, statistical results, and interpretations used in the paper were reviewed by the authors.

\bibliographystyle{ACM-Reference-Format}
\bibliography{biblio}

@article{technologies12090157,
author = {Alotaibi, Albandari and Alatawi, Hanan and Binnouh, Aseel and Duwayriat, Lamaa and Alhmiedat, Tareq and Alia, Osama Moh’d},
title = {Deep Learning-Based Vision Systems for Robot Semantic Navigation: An Experimental Study},
journal = {Technologies},
volume = {12},
year = {2024},
number = {9},
article-number = {157},
url = {https://www.mdpi.com/2227-7080/12/9/157},
issn = {2227-7080},
doi = {10.3390/technologies12090157}
}

@article{9764831,
author = {Tang, Yang and Zhao, Chaoqiang and Wang, Jianrui and Zhang, Chongzhen and Sun, Qiyu and Zheng, Wei Xing and Du, Wenli and Qian, Feng and Kurths, Jürgen},
journal = {IEEE Transactions on Neural Networks and Learning Systems}, 
title = {Perception and Navigation in Autonomous Systems in the Era of Learning: A Survey}, 
year = {2023},
volume = {34},
number = {12},
pages = {9604--9624},
doi = {10.1109/TNNLS.2022.3167688}
}

@inproceedings{yousaf2026vlmUnderwater,
author = {Yousaf, Muhammad and Arrieta, Aitor and Ali, Shaukat and Arcaini, Paolo and Wang, Shuai},
booktitle = {2026 IEEE International Conference on Software Testing, Verification and Validation (ICST)}, 
title = {Assessing Vision–Language Models for Perception in Autonomous Underwater Robotic Software}, 
year = {2026},
volume = {},
number = {},
pages = {64--75},
doi = {10.1109/ICST69053.2026.00021}
}

@inproceedings{wu2026vlmTestingAMR,
author = {Wu, Jiahui and Lu, Chengjie and Arrieta, Aitor and Ali, Shaukat and Peyrucain, Thomas},
title = {Vision Language Model-based Testing of Industrial Autonomous Mobile Robots},
year = {2026},
isbn = {9798400724268},
publisher = {Association for Computing Machinery},
address = {New York, NY, USA},
url = {https://doi.org/10.1145/3786583.3786920},
doi = {10.1145/3786583.3786920},
booktitle = {Proceedings of the IEEE/ACM 48th International Conference on Software Engineering: Software Engineering in Practice},
pages = {831--841},
numpages = {11},
location = {Rio de Janeiro, Brazil},
series = {ICSE-SEIP '26}
}

@inproceedings{10.1145/3805760.3814910,
author = {Saurabh, Prasun and Valle, Pablo and Arrieta, Aitor and Ali, Shaukat and Arcaini, Paolo},
title = {{VISOR}: {A} Vision-Language Model-Based Test Oracle for Testing Robots},
year = {2026},
isbn = {9798400726019},
publisher = {Association for Computing Machinery},
address = {New York, NY, USA},
url = {https://doi.org/10.1145/3805760.3814910},
doi = {10.1145/3805760.3814910},
booktitle = {Proceedings of the 3rd ACM International Conference on AI-Powered Software},
pages = {197--207},
numpages = {11},
location = {Montreal, QC, Canada},
series = {AIware '26}
}

@article{DBLP:journals/corr/abs-2303-03378,
author = {Danny Driess and
Fei Xia and
Mehdi S. M. Sajjadi and
Corey Lynch and
Aakanksha Chowdhery and
Brian Ichter and
Ayzaan Wahid and
Jonathan Tompson and
Quan Vuong and
Tianhe Yu and
Wenlong Huang and
Yevgen Chebotar and
Pierre Sermanet and
Daniel Duckworth and
Sergey Levine and
Vincent Vanhoucke and
Karol Hausman and
Marc Toussaint and
Klaus Greff and
Andy Zeng and
Igor Mordatch and
Pete Florence},
title = {{PaLM-E: An} Embodied Multimodal Language Model},
journal = {CoRR},
volume = {abs/2303.03378},
year = {2023},
url = {https://doi.org/10.48550/arXiv.2303.03378},
doi = {10.48550/ARXIV.2303.03378},
eprinttype = {arXiv},
eprint = {2303.03378}
}

@article{DBLP:journals/corr/abs-2307-15818,
author = {Anthony Brohan and
Noah Brown and
Justice Carbajal and
Yevgen Chebotar and
Xi Chen and
Krzysztof Choromanski and
Tianli Ding and
Danny Driess and
Avinava Dubey and
Chelsea Finn and
Pete Florence and
Chuyuan Fu and
Montse Gonzalez Arenas and
Keerthana Gopalakrishnan and
Kehang Han and
Karol Hausman and
Alexander Herzog and
Jasmine Hsu and
Brian Ichter and
Alex Irpan and
{Nikhil J.} Joshi and
Ryan Julian and
Dmitry Kalashnikov and
Yuheng Kuang and
Isabel Leal and
Lisa Lee and
Tsang{-}Wei Edward Lee and
Sergey Levine and
Yao Lu and
Henryk Michalewski and
Igor Mordatch and
Karl Pertsch and
Kanishka Rao and
Krista Reymann and
{Michael S.} Ryoo and
Grecia Salazar and
Pannag Sanketi and
Pierre Sermanet and
Jaspiar Singh and
Anikait Singh and
Radu Soricut and
{Huong T.} Tran and
Vincent Vanhoucke and
Quan Vuong and
Ayzaan Wahid and
Stefan Welker and
Paul Wohlhart and
Jialin Wu and
Fei Xia and
Ted Xiao and
Peng Xu and
Sichun Xu and
Tianhe Yu and
Brianna Zitkovich},
title = {{RT-2:} Vision-Language-Action Models Transfer Web Knowledge to Robotic
Control},
journal = {CoRR},
volume = {abs/2307.15818},
year = {2023},
url = {https://doi.org/10.48550/arXiv.2307.15818},
doi = {10.48550/ARXIV.2307.15818},
eprinttype = {arXiv},
eprint = {2307.15818}
}

@inproceedings{Zhang-RSS-24, 
author = {Jiazhao Zhang AND Kunyu Wang AND Rongtao Xu AND Gengze Zhou AND Yicong Hong AND Xiaomeng Fang AND Qi Wu AND Zhizheng Zhang AND He Wang}, 
title = {{NaVid: Video-based VLM Plans the Next Step for Vision-and-Language Navigation}}, 
booktitle = {Proceedings of Robotics: Science and Systems}, 
year = {2024}, 
address = {Delft, Netherlands}, 
month = {July}, 
doi = {10.15607/RSS.2024.XX.079} 
}

@inproceedings{DBLP:conf/iclr/HendrycksG17,
author = {Dan Hendrycks and Kevin Gimpel},
title = {A Baseline for Detecting Misclassified and Out-of-Distribution Examples in Neural Networks},
booktitle = {5th International Conference on Learning Representations, {ICLR} 2017, Toulon, France, April 24-26, 2017, Conference Track Proceedings},
publisher = {OpenReview.net},
year = {2017},
url = {https://openreview.net/forum?id=Hkg4TI9xl}
}

@inproceedings{3305381.3305518,
author = {Guo, Chuan and Pleiss, Geoff and Sun, Yu and Weinberger, Kilian Q.},
title = {On calibration of modern neural networks},
year = {2017},
publisher = {JMLR.org},
booktitle = {Proceedings of the 34th International Conference on Machine Learning - Volume 70},
pages = {1321--1330},
numpages = {10},
location = {Sydney, NSW, Australia},
series = {ICML'17}
}

@article{DBLP:journals/corr/abs-2508-07611,
author = {Zifan Wang and Xun Yang and Jianzhuang Zhao and Jiaming Zhou and Teli Ma and Ziyao Gao and Arash Ajoudani and Junwei Liang},
title = {End-to-End Humanoid Robot Safe and Comfortable Locomotion Policy},
journal = {CoRR},
volume = {abs/2508.07611},
year = {2025},
url = {https://doi.org/10.48550/arXiv.2508.07611},
doi = {10.48550/ARXIV.2508.07611},
eprinttype = {arXiv},
eprint = {2508.07611}
}

@article{10.3389/frobt.2026.1870024,
author = {Peña-Narvaez, Juan Diego and Pérez-Rodríguez, Rodrigo and Aguado, Esther and Moreno, Francisco Miguel and Rico, Francisco Martín }, 
title = {Predictive vision-language monitoring for proactive safety in robot task execution}, 
journal = {Frontiers in Robotics and AI}, 
volume = {Volume 13 - 2026},
year = {2026},
url = {https://www.frontiersin.org/journals/robotics-and-ai/articles/10.3389/frobt.2026.1870024},
doi = {10.3389/frobt.2026.1870024},
issn = {2296-9144},
}

@inproceedings{NEURIPS2025_392d0d05,
author = {Gu, Qiao and Ju, Yuanliang and Sun, Shengxiang and Gilitschenski, Igor and Nishimura, Haruki and Itkina, Masha and Shkurti, Florian},
booktitle = {Advances in Neural Information Processing Systems},
doi = {10.52202/085713-1337},
editor = {D. Belgrave and C. Zhang and H. Lin and R. Pascanu and P. Koniusz and M. Ghassemi and N. Chen},
pages = {40041--40076},
publisher = {Curran Associates, Inc.},
title = {{SAFE: Multitask} Failure Detection for Vision-Language-Action Models},
url = {https://proceedings.neurips.cc/paper_files/paper/2025/file/392d0d05e2f514063e6ce6f8b370834c-Paper-Conference.pdf},
volume = {38, Main Conference},
year = {2025}
}

@inproceedings{11444867,
author = {Lafon, Marc and Karmim, Yannis and Silva-Rodríguez, Julio and Couairon, Paul and Rambour, Clément and Fournier-Sniehotta, Raphaël and Ben Ayed, Ismail and Dolz, Jose and Thome, Nicolas},
booktitle = {2025 IEEE/CVF International Conference on Computer Vision (ICCV)}, 
title = {{ViLU: Learning} Vision-Language Uncertainties for Failure Prediction}, 
year = {2025},
volume = {},
number = {},
pages = {17807--17817},
doi = {10.1109/ICCV51701.2025.01654}
}

@article{s26103140,
author = {Ghaleb, Atef M. and Allahloh, Ali S. and Mejjaouli, Sobhi and Ali, Mohammed A. H. and Al-Shayea, Adel},
title = {Uncertainty-Calibrated Safety Gating for Vision–Language– Action Manipulation Under Domain Shift: Reliability Gains and Intervention–Efficiency Trade-Offs},
journal = {Sensors},
volume = {26},
year = {2026},
number = {10},
article-number = {3140},
url = {https://www.mdpi.com/1424-8220/26/10/3140},
PubMedID = {42197948},
issn = {1424-8220},
doi = {10.3390/s26103140}
}

@article{YANG2026131742,
title = {{FPC-VLA: A} vision-language-action framework with a supervisor for failure prediction and correction},
journal = {Expert Systems with Applications},
volume = {316},
pages = {131742},
year = {2026},
issn = {0957-4174},
doi = {https://doi.org/10.1016/j.eswa.2026.131742},
url = {https://www.sciencedirect.com/science/article/pii/S095741742600655X},
author = {Yifan Yang and Zhixiang Duan and Tianshi Xie and Fuyu Cao and Pinxi Shen and Peili Song and Chenyang Zhao and Piaopiao Jin and Guokang Sun and Shaoqing Xu and Yangwei You and Jingtai Liu}
}

@article{10.1108/IR-02-2025-0074,
title = {Large language and vision-language models for robot: safety challenges, mitigation strategies and future directions},
journal = {Industrial Robot},
volume = {53},
number = {1},
pages = {88--101},
year = {2026},
issn = {0143-991X},
doi = {https://doi.org/10.1108/IR-02-2025-0074},
url = {https://www.sciencedirect.com/science/article/pii/S0143991X2500056X},
author = {Xiangyu Hu and Zhenyu Xu}
}

@article{LIU2025100943,
title = {Vision language model-enhanced embodied intelligence for digital twin-assisted human-robot collaborative assembly},
journal = {Journal of Industrial Information Integration},
volume = {48},
pages = {100943},
year = {2025},
issn = {2452-414X},
doi = {https://doi.org/10.1016/j.jii.2025.100943},
url = {https://www.sciencedirect.com/science/article/pii/S2452414X25001669},
author = {Changchun Liu and Dunbing Tang and Haihua Zhu and Zequn Zhang and Liping Wang and Yi Zhang}
}

@misc{palTiagoOmniHandbook,
author = {{PAL Robotics}},
title = {{TIAGo OMNI Base Handbook}},
howpublished = {\url{https://docs.pal-robotics.com/tiago-omni-base/handbook.html}},
note = {Accessed: 2026}
}

@book{ling2016university,
title = {University Physics Volume 1},
author = {Ling, Samuel J. and Sanny, Jeff and Moebs, William},
year = {2016},
publisher = {OpenStax},
address = {Houston, TX},
isbn = {1938168275, 9781938168277},
url = {https://openstax.org/books/university-physics-volume-1/pages/3-4-motion-with-constant-acceleration},
note = {Section 3.4: Motion with Constant Acceleration, Equation 3.12}
}

@misc{farooq2026vlmrepo,
author = {Muhammad Farooq and Aitor Arrieta and Paolo Arcaini and Shaukat Ali and Thomas Peyrucain},
title = {Supplementary Material and Replication Package for the paper ``Assessing Vision-Language Models for Early Prediction of Unsafe Situations in Robotics''},
year = {2026},
howpublished = {\url{https://github.com/mufar6566/VLM-Early-Warning-Robot-Safety}},
note = {GitHub repository}
}

@misc{rosBumperMsg,
author = {{ROS}},
title = {ca\_msgs/Bumper Message Documentation},
howpublished = {\url{https://docs.ros.org/jade/api/ca_msgs/html/msg/Bumper.html}},
note = {Accessed: 2026-10-20}
}

@article{10.3389/fpace.2026.1937868,
author = {Ma, Sizhe and Flanigan, Katherine A. and Bergés, Mario }, 
title = {Benchmarking time-series foundation models for multitask predictive maintenance in digital twins}, 
journal = {Frontiers in Aerospace Engineering}, 
volume = {Volume 5 - 2026},
year = {2026},
url = {https://www.frontiersin.org/journals/aerospace-engineering/articles/10.3389/fpace.2026.1937868},
doi = {10.3389/fpace.2026.1937868},
issn = {2813-2831},
}

\end{document}